\documentclass[onecolumn,11pt]{article}
\usepackage[top=1in, bottom=1in, left=1in, right=1in]{geometry}
\usepackage{amsmath,amsfonts,amscd,amssymb}
\usepackage{graphicx}
\usepackage{epstopdf}
\usepackage{overpic}
\usepackage{cancel}
\usepackage{rotating}
\usepackage{url}
\usepackage{caption}
\usepackage{color}
\usepackage{rotating}
\usepackage{multirow}
\usepackage{wrapfig}
\usepackage{mathtools}
\usepackage{subeqnarray}
\usepackage{setspace}
\usepackage{palatino} 
\usepackage{hyperref}
\usepackage[numbers,sort&compress]{natbib}
\usepackage[mathscr]{euscript}

\usepackage{url}

\usepackage{microtype}
\usepackage{pict2e}
\usepackage{graphicx}
\usepackage{booktabs} %
\usepackage{hyperref}

\usepackage{caption}
\usepackage{natbib}

\usepackage{amsmath,amssymb,amsfonts,amsthm}

\usepackage{xcolor}
\usepackage{mathtools}
\usepackage{dsfont}
\usepackage{hyperref}
\usepackage{bm}
\usepackage{nicefrac}
\usepackage{wrapfig}
\usepackage{lipsum}

\usepackage{algorithm,algorithmic}
\usepackage[linesnumbered, ruled,vlined,algo2e]{algorithm2e}

\newcounter{assumption}%
\renewcommand{\theassumption}{\arabic{assumption}}

\newcommand*\Ep[2]{\mathbb{E}_{#1}\left[#2\right]}

\newcommand*\lrb[1]{\left[#1\right]}

\newcommand*\lrp[1]{\left(#1\right)}
\newcommand*\lrn[1]{\left\|#1\right\|}

\newcommand{\real}{\ensuremath{\mathbb{R}}}

\newcommand{\param}{\theta}

\newcommand{\prior}{p_\param}

\usepackage{multirow}
\usepackage{caption}
\usepackage{subcaption}

\newcommand{\ve}[1]{\mathbf{#1}}

\usepackage[bottom,flushmargin,hang,multiple]{footmisc}
\usepackage{lipsum}

\definecolor{header1}{cmyk}{0,0,0,1}

\newcommand{\Xv}{\mathbf{X}}
\newcommand{\Zv}{\mathbf{Z}}

\newcommand{\fv}{\mathbf{f}}
\newcommand{\gv}{\mathbf{g}}

\newcommand{\xv}{\mathbf{x}}
\newcommand{\zv}{\mathbf{z}}
\newcommand{\thetav}{\mathbf{\theta}}
\newcommand{\Thetav}{\boldsymbol{\Theta}}

\newcommand{\Xiv}{\boldsymbol{\Xi}}
\newcommand{\xiv}{\boldsymbol{\xi}}

\newcommand\Rb{\mathbb{R}}

\title{\LARGE{\vspace{-.55in}\textbf{Simultaneous discovery of coordinates and parsimonious dynamics: SINDy autoencoders}}\vspace{-.175in}}
\title{\vspace{-.55in}{\fontsize{16}{16}\selectfont \textbf{Bayesian autoencoders for data-driven discovery of coordinates, governing equations and fundamental constants}}\vspace{-.15in}}

\author{\normalsize{L. Mars Gao$^{1}$, J. Nathan Kutz$^2$\thanks{Corresponding author}}\\
\footnotesize{$^1$ Paul G. Allen School of Computer Science \& Engineering, University of Washington, Seattle, WA}\\
\footnotesize{$^2$ Department of Applied Mathematics and Electrical and Computer Engineering, University of Washington, Seattle, WA\vspace{-.2in}}
}

\date{}
\begin{document}
\maketitle
\vspace{-.2in}
\begin{abstract}
%
Recent progress in autoencoder-based sparse identification of nonlinear dynamics (SINDy) under $\ell_1$ constraints allows joint discoveries of governing equations and latent coordinate systems from spatio-temporal data, including simulated video frames. 
%
%
However, it is challenging for  $\ell_1$-based sparse inference to perform correct identification for real data due to the noisy measurements and often limited sample sizes. 
To address the data-driven discovery of physics in the low-data and high-noise regimes, we propose Bayesian SINDy autoencoders, which incorporate a hierarchical Bayesian sparsifying prior: Spike-and-slab Gaussian Lasso. Bayesian SINDy autoencoder enables the joint discovery of governing equations and coordinate systems with a theoretically guaranteed uncertainty estimate. 
%
%
To resolve the challenging computational tractability of the Bayesian hierarchical setting, we adapt an adaptive empirical Bayesian method with Stochatic gradient Langevin dynamics (SGLD) which gives a computationally tractable way of Bayesian posterior sampling within our framework.
Bayesian SINDy autoencoder achieves better physics discovery with lower data and fewer training epochs, along with valid uncertainty quantification suggested by the experimental studies. 
The Bayesian SINDy autoencoder can be applied to real video data, with accurate physics discovery which correctly identifies the governing equation and provides a close estimate for standard physics constants like gravity $g$, for example, in videos of a pendulum. 

\end{abstract}

\section{Introduction}


Calculus-based models fundamentally relate the rates of change of quantities of interest in time and space through differential and partial differential equations.
From population models to turbulence, physics and engineering principles are rooted in such governing equations.
In the modern era of big data, there is a growing demand to transform rich spatio-temporal data into descriptive physical models in an automated, data-driven fashion. 
Video data, for example, contain optical snapshots of an observed dynamical system described by some specific governing equations. 
To understand the underlying physics, it is important not only to identify the equations, but to discover dependent state variables (sparse representations) directly from the video frames.  To discover such an underlying sparse representation, it is essential to find a correct coordinate transformation (latent space or manifold) to compress the data into a low-dimensional space. Principal Component Analysis (PCA) is often applied to obtain a low-dimensional subspace with linearity constraints. For nonlinear transformations,  Autoencoders with neural networks are frequently applied as a nonlinear extension of PCA \cite{hinton2006reducing,bao2020regularized}. To identify the governing equation given the latent sparse representations, one can apply data-driven methods via sparse regression~\cite{kutz2016dynamic,schaeffer2017learning,rudy2017data}. The sparse regression-based model discovery enables a computationally efficient way of identifying governing equation with convergence guarantees~\citep{zhang2019convergence}. 
In this case of physics discovery from videos, Champion et al.~\cite{champion2019data} propose SINDy autoencoders which can jointly discover on coordinates and equations for synthetic video data. More recently, Chen et al.~\citep{chen2021discovering} introduce the Neural State Variables learning from video data, which enables automated discovery of the latent state variables from the underlying dynamical system.~\citep{chen2021discovering}. 

The automated discovery of coordinates and equations in real video is significantly more challenging compared to these prior works~\cite{champion2019discovery,chen2021discovering}.  First, the temporal derivatives are not directly available from the video data.
To resolve the missing temporal derivatives, one has to find an approximation for the temporal derivatives from discrete video frames, which is frequently numerically unstable and will inject a high level of noise. Indeed, lighting effects alone can greatly compromise derivate estimates.
Second, the sample size of the real video dataset may not be sufficiently large in comparison with synthetic video~\citep{champion2019data}. 
These pragmatic constraints hinder the real video data from having a statistically sufficient sample size for learning. 
Thus automated discovery from real video data is much more challenging due to being in the low-data and high-noise limit. 

In this low-data and high-noise regime, Bayesian sparse regression methods generally have significant advantages from both a theoretical and practical perspective~\citep{mitchell1988bayesian,george1997approaches,scott2014predicting}. The spike-and-slab prior~\citep{ishwaran2005spike} (e.g. Bernoulli-Gaussian, Bernoulli-Laplace~\citep{amini2012analog}) with hierarchical Bayesian settings has a proven success for both sparse variable selection and uncertainty quantification~\citep{rovckova2014emvs}. 
In data-driven model discovery, Bayesian SINDy~\citep{hirsh2022sparsifying} with the spike-and-slab prior has also shown an advantage for correct model identification in the case study of Lynx-hare population~\citep{hewitt1921conservation} under the very low-data limit. 
However, a significant limitation of Bayesian methods comes from its high computational cost that hinders both speed and scalability. 
The MCMC-based hierarchical Bayesian model sampling requires a considerably long run because the underlying stochastic binary search grows exponentially with the number of parameters. Additionally, it is very costly to compute the full gradient of the entire video dataset for the MCMC sampling. 
Therefore, even if the Bayesian methods are much more powerful in the low-data regime, it is highly nontrivial to design a feasible Bayesian solution for the discovery of governing equations and coordinate systems given the computationally intractability. 

In this paper, we propose Bayesian SINDy autoencoders that extends SINDy Autoencoder \cite{champion2019data} into a Bayesian learning framework.  Bayesian SINDy autoencoders can perform a joint discovery for governing equations and coordinate systems in a computationally tractable manner.  Specifically, we apply the Spike-and-slab Gaussian-Laplace (SSGL) prior to the intermediate SINDy module to accelerate the sparse identification process of governing equation discovery. 
To resolve the computational burden arising from the hierarchical Bayesian model, we consolidate the Bayesian sampling procedure via Stochastic Gradient Langevin Dynamics (SGLD) with an adaptive empirical Bayesian variable selection method using Expectation–maximization. 
Instead of computing the full batch gradient, SGLD evaluates mini-batch gradients with injected random Gaussian noise, which is theoretically valid to generate Langevin-based proposal distribution~\citep{welling2011bayesian,chandra2022revisiting}. The mini-batch gradient learning naturally fits into the training of deep neural networks and relaxes the scalability issue at the same time. 
On the other hand, the adaptive Bayesian Expectation–maximization variable selection (EMVS) performs variable selection by optimizing the latent inclusion probability of each variable, which avoids the previously required lengthy binary stochastic search. 
Adapting these two ideas into the SINDy module, we can simultaneously perform an accelerated governing equation identification under the low-data, high-noise limit with a trustworthy uncertainty quantification through the power of Bayesian estimation. 

In our numerical experiments, Bayesian SINDy autoencoders achieve accelerated discovery in governing equations and coordinate systems under the SSGL prior. 
In addition to the model discovery in synthetic datasets, from a real video data on a single moving pendulum, we obtain correct governing equation discovery with a close estimate of the gravity constant $\hat{g}=-9.876$ with only 390 data snapshots. 
With precise understanding of the underlying physics, Bayesian SINDy autoencoder enables explainable, trustworthy, and robust video predictions. 
In summary, the contribution of this paper is threefold: 
\begin{enumerate}
    \item We propose the Bayesian SINDy autoencoder with Spike-and-slab Gaussian-Laplace prior to accelerated sparse inference under low-data and high-noise environments. 
    \item We utilize Stochastic Gradient Langevin Dynamics to perform posterior sampling in Bayesian SINDy autoencoder with valid uncertainty estimations. 
    \item We conduct extensive experiments with successful joint discoveries on coordinate systems and governing equations for both synthetic and real datasets. Remarkably, we achieve a correct physics discovery with a $14$ seconds recording on our experiment on a pendulum. 
\end{enumerate}

\section{Background}\label{sec:background}

The current work is built upon two primary mathematical innovations:  (i) sparse regression used in the SINDy algorithm, including learning latent representations, and (ii) Bayesian learning with sparse priors. A quick review of each is given in order to better inform the reader how they are combined into our Bayesian SINDy autoencoder framework.

\subsection{Sparse identification of nonlinear dynamics}

We review the {\em sparse identification of nonlinear dynamics} (SINDy) \cite{brunton2016discovering} algorithm, which utilizes sparse regression to identify the latent dynamical system from snapshot data. 
SINDy takes snapshot data $\xv(t) \in \Rb^n$ and aims to discover the underlying dynamical system
%
\begin{equation}
  \dot{\ve{x}}(t) = \fv(\xv(t)).
    \label{eq:dynamical_system_x}
\end{equation}
The snapshots are collected by measurements at time $t\in [t_1, t_m]$, and the function $\fv$ characterizes the dynamics. Assuming the temporal derivatives of the snapshot are available from data, SINDy forms data matrices in the following way: 
\begin{equation}
  \Xv = \left(\begin{array}{cccc}
    \xv (t_1) \\
    \xv (t_2) \\
    \vdots \\
    \xv (t_m) \\
  \end{array}\right), \quad
  \dot{\Xv} = \left(\begin{array}{cccc}
    \dot{\xv}_1(t_1) \\
    \dot{\xv}_1(t_2) \\
    \vdots \\
    \dot{\xv}_1(t_m) \\
  \end{array}\right), \nonumber
\end{equation}
with $\Xv,\dot{\Xv} \in \Rb^{m \times n}$. The candidate function library is constructed by $p$ candidate model term $\theta_j$'s that $\Thetav(\Xv) = [\thetav_1(\Xv) \cdots \thetav_p(\Xv)] \in \Rb^{m\times p}$. 
A common choice of candidate functions are polynomials in $\xv$ targeting common canonical models of dynamical systems~\cite{guckenheimer2013nonlinear}. The Fourier library is also very common with $\sin(\cdot)$ and $\cos(\cdot)$ terms. 
In summary, we build a model between $\ve{X}$ and $\dot{\ve{X}}$ that
\begin{equation}
  \dot{\Xv} = \Thetav(\Xv)\Xiv \nonumber
\end{equation}
where the unknown matrix $\Xiv = (\xiv_1\ \xiv_2\ \cdots\ \xiv_n)\in \Rb^{p\times n}$ is the set of coefficients. The sparse inference on $\Xiv$ enables sparse identification of the dynamical system $\fv$. 
For high-dimensional systems, the goal is to jointly identify a low-dimensional state $\zv=\varphi(\xv)$ with dynamics $\dot{\zv} = \gv(\zv)$. The standard SINDy approach uses a sequentially thresholded least squares algorithm to perform sparse inference \cite{brunton2016discovering}, which is a proxy for $\ell_0$ optimization~\cite{zheng2018unified} with convergence guarantees~\cite{zhang2019convergence}.  
Small et al~\cite{small2002modeling} and
Yao and Bollt~\cite{yao2007modeling} previously formulated the dynamical system identification without sparsity constraints.   
These methods provide a computationally efficient counterpart to other model discovery frameworks~\cite{schmidt2009distilling,wang2011predicting}. 

In an alternative approach, Hirsh et al.~\citep{hirsh2022sparsifying} utilize Bayesian sparse regression techniques in model discovery, which have improved the
robustness of SINDy under high-noise and low-data settings. Bayesian sparse inference models the sparsity with various probabilistic priors $p(\Xi)$ like the Spike-and-slab~\citep{ishwaran2005spike}, regularized horseshoe~\citep{carvalho2009handling,carvalho2010horseshoe}, Laplace priors~\citep{park2008bayesian,tibshirani1996regression} and so on. After defining the Bayesian likelihood function with sparsifying priors, Bayesian SINDy generates posterior distribution via the No-U-Turn MCMC sampler~\citep{hoffman2014no}. Since MCMC for sparse inference can be extremely computationally demanding,~\citep{rovckova2014emvs,rovckova2018spike} propose coordinate descent sparse inference methods via Spike-and-slab prior, targeting the mode detection. As an approximation to Bayesian inference, Fasel et al.~\citep{fasel2022ensemble} combine ensembling techniques via bootstrapping to perform uncertainty estimation and accelerated variable selection for system identification.  

SINDy has been widely applied to model discovery for many scientific scenarios including fluid dynamics~\cite{loiseau2018constrained,loiseau2018sparse,loiseau2020data,guan2021sparse,deng2021galerkin,callaham2022empirical}, nonlinear optics~\cite{sorokina2016sparse}, turbulence
closures~\citep{beetham2020formulating,beetham2021sparse,schmelzer2020discovery}, ocean closures~\citep{zanna2020data}, chemical reaction~\cite{hoffmann2019reactive},  plasma dynamics~\cite{dam2017sparse,alves2022data,kaptanoglu2021physics}, structural modeling~\cite{lai2019sparse}, and for model predictive control~\cite{kaiser2018sparse}. 
There are many extensions of SINDy, including the identification of partial differential equations~\cite{rudy2017data,schaeffer2017learning},  multiscale physics~\cite{champion2019discovery}, parametrically dependent dynamical models~\cite{rudy2019data},  time-dependent PDEs~\citep{chen2021robust}, switching dynamical systems~\cite{mangan2019model}, rational function nonlinearities~\cite{mangan2016inferring,kaheman2020sindy}, control inputs~\citep{kaiser2018sparse}, constraints on symmetries~\citep{loiseau2018constrained}, control for stability~\citep{kaptanoglu2021promoting}, control for robustness~\citep{alves2022data,schaeffer2017sparse,reinbold2020using,reinbold2021robust,messenger2021weak,messenger2021weak}, stochastic dynamical systems~\citep{boninsegna2018sparse,callaham2021nonlinear}, and multidimensional approximation on tensors~\citep{gelss2019multidimensional}.

A related and important extension to the SINDy framework is  the
SINDy autoencoder~\citep{champion2019data}, which embeds SINDy into the training process of deep autoencoders. The SINDy autoencoder achieves remarkable performances on high-dimensional synthetic data to jointly discover coordinate systems and governing equations. 
Due to over-parametrization  and non-convexity, deep neural networks do not generally have explainable guarantees for inference and predictions. 
Dynamical system learning and forecasting is typically an extrapolatory problem by nature.  
Therefore, the interpretability of models is very important to understand.  
In this case, SINDy autoencoder is satisfactory due to the transparency of the learning process.  Specifically, the encoder and decoder focus on the specific task of learning only a coordinate transformation, with the SINDy layer  targeting the inference of latent dynamical systems. 
The SINDy autoencoder can not only identify the governing equation from high-dimensional data but can also perform trustworthy predictions based on future dynamics.  The limitations of the SINDy autoencoder are exactly what our Bayesian framework addresses.

\subsection{Bayesian and sparse deep learning}
Bayesian deep learning achieves outstanding success in various machine learning tasks like computer vision~\citep{gustafsson2020evaluating}, physics-informed modeling~\citep{yang2021b}, and complex dynamical system control~\citep{watter2015embed}. From previous works, the main advantages of Bayesian deep learning come from two parts.  First, the Bayesian framework can unify uncertainty quantification in deep learning. This includes the uncertainty from the neural network parameters, task-specific parameters, and exchanging information~\citep{wang2020survey,abdar2021review}, which is applicable to computer vision~\citep{kendall2017uncertainties}, spatiotemporal forecasting~\citep{wu2021quantifying}, weather forecasting~\citep{wang2019deep}, and so on~\citep{zhu2018bayesian}. 
Second, the Bayesian framework allows theoretically grounded ensemble neural networks via model averaging~\citep{wilson2020bayesian}. Bayesian neural networks apply weights distribution to neurons~\citep{blundell2015weight}.  
It is important to avoid computational requirements from Bayesian when applying it to deep learning. 
To perform Bayesian sampling in deep learning, Welling and Teh establish Stochastic Gradient Langevin Dynamics as an MCMC sampler in mini-batch settings~\citep{welling2011bayesian}. Neal introduces MCMC via Hamiltonian dynamics~\citep{neal2011mcmc}, and Chen et al. extend into deep learning via Stochastic gradient Hamiltonian Monte Carlo (SGHMC)~\citep{chen2014stochastic,ma2015complete}. Other techniques include Nosé–Hoover thermostat~\citep{ding2014bayesian}, replica-exchange SGHMC~\citep{deng2020non}, cyclic SGLD~\citep{zhang2019cyclical}, contour SGLD~\citep{deng2020contour}, preconditioned SGLD~\citep{wang2021bayesian}, and adaptively weighted SGLD~\citep{deng2022adaptively}. 
Variational methods could also help to perform approximate inference for Bayesian deep learning~\citep{hernandez2015probabilistic,blundell2015weight}. 

Sparse deep learning is of emerging interest due to its lowered computational cost and improved interpretability. 
The discussion on sparse neural networks dates back decades~\citep{elizondao1995non,morgan2008differential}. 
Glorot et al. use Rectifier Neurons with sparsity constraints to obtain sparse representations~\citep{glorot2011deep}. 
Liu et al. extend to Convolutional neural networks under sparse settings~\citep{liu2015sparse}. Then, several follow-up works aimed to train sparse neural networks efficiently~\citep{louizos2017learning,srinivas2017training}. 
By adapting Bayesian sparse inference methods, the sparse inference process could be accelerated with valid uncertainty quantification via various prior options~\citep{sun2022learning,deng2019adaptive,wang2020uncertainty}.
Variational methods are also applied for efficient Bayesian inference~\citep{bai2020efficient}. 
There have also been theoretical discussion of Bayesian sparse deep learning~\citep{sun2021consistent,polson2018posterior,sun2021sparse}.  
Overall, sparse deep learning has the potential to make computationally tractable the construction of a Bayesian model.

\section{Bayesian SINDy Autoencoders}

This section presents the SINDy autoencoder with a Bayesian learning process that incorporates sparsifying priors and posterior sampling. We first introduce the Sparse identification of nonlinear dynamics (SINDy) autoencoder in Sec.~\ref{sec:sindy}. Then, we propose a Bayesian learning framework that includes SINDy autoencoder in the likelihood function and specifies various setups for different sparsifying priors. Finally, we discuss Stochastic Gradient Langevin Dynamics (SGLD) to generate posterior samples from the Bayesian learning model. 

\subsection{Likelihood setting for Bayesian SINDy Autoencoders} \label{sec:sindy}

\begin{figure*}
\vspace{-.1in}
 \centering
\begin{overpic}[width=\textwidth]{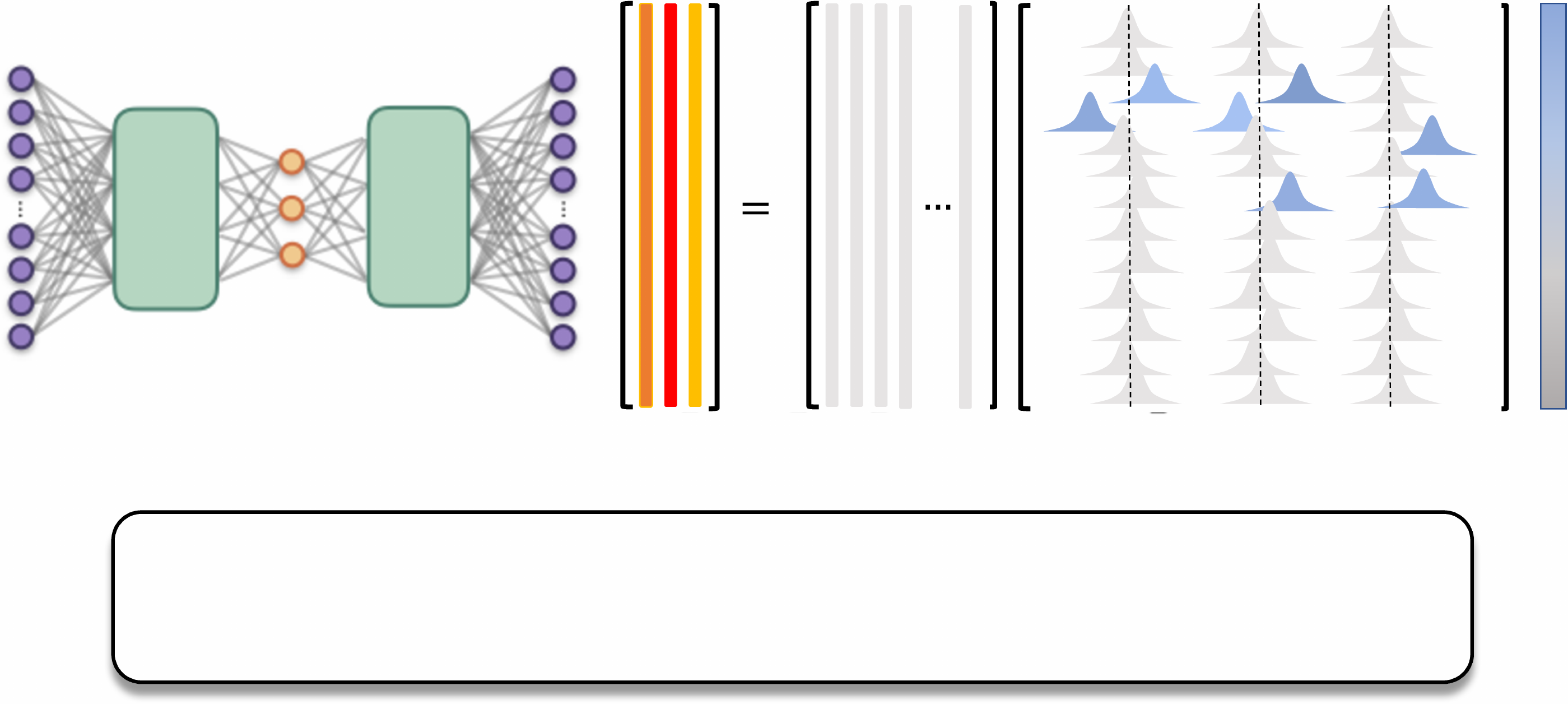}
  \put(-5.5,46){(a)}
  \put(34.5,46){(b)}
  \put(-0.5,20){\large $\xv(t)$}
  \put(16.5,20){\large $\zv(t)$}
  \put(33.5,20){\large $\hat{\xv}(t)$}
  \put(10,31){\large $f$}
  \put(26,31.5){\large $g$}
  \put(39.5,45.5){\small $\dot{z}_1$}
  \put(41.3,45.5){\small $\dot{z}_2$}
  \put(43.5,45.5){\small $\dot{z}_3$}
  \put(52,45.5){\tiny $1$}
  \put(53,45.5){\tiny $z_1$}
  \put(54.5,45.5){\tiny $z_2$}
  \put(56,45.5){\tiny $z_3$}
  \put(60.6,45.5){\tiny $z_3^2$}
  \put(71,45.5){\small $\xi_1$}
  \put(79.5,45.5){\small $\xi_2$}
  \put(87.5,45.5){\small $\xi_3$}
  \put(42,16){\large $\dot{\Zv}$}
  \put(55,16){\large $\Thetav(\Zv)$}
  \put(79,16){\large $\Xiv$}
  \put(55,13){$\dot{\zv}_i = \nabla_\xv f(\xv_i)\dot{\xv}_i$}
  \put(75,13){$\Thetav(\zv_i^T) = \Thetav(f(\xv_i)^T)$}
  \put(99.82,18.46){\tiny \textit{-} $0.0$}
  \put(99.82,24.91){\tiny \textit{-} $0.25$}
  \put(99.82,31.36){\tiny \textit{-} $0.5$}
  \put(99.82,37.81){\tiny \textit{-} $0.75$}
  \put(99.82,44.28){\tiny \textit{-} $1.0$}
  \put(11,9){\small$\underbrace{\left\| \xv - g(\zv)\right\|_2^2} + \underbrace{\lambda_1 \left\|\dot{\xv} - \left(\nabla_\zv g(\zv)\right)\left(\Thetav(\zv^T)\Xiv\right)\right\|_2^2} + \underbrace{\lambda_2 \left\|\left(\nabla_\xv\zv\right)\dot{\xv} - \Thetav(\zv^T)\Xiv\right\|_2^2} + \underbrace{\lambda_3 \log p(\Xiv)}$}
  \put(8.5,4.2){\small reconstruction loss}
  \put(31.5,4.2){\small SINDy loss in $\dot{\xv}$}
  \put(59.5,4.2){\small SINDy loss in $\dot{\zv}$}
  \put(80,3.7){\small \parbox{6em}{\centering SINDy\\prior}}
\end{overpic}
\vspace{3mm}
\caption{General structure of the Bayesian SINDy autoencoder architecture. (a) An autoencoder architecture is used to discover intrinsic coordinates $\zv$ from high-dimensional input data $\xv$. The encoder $f(\xv)$ transforms the high-dimensional input into the a low-dimensional space $\zv$, and the decoder $g(\zv)$ reconstructs $\xv$ from the low-dimensional subspace. (b) A Bayesian SINDy model infers the underlying dynamics of the latent dimension with uncertainty estimation. The active terms are identified by $\Xiv$. The color represents the inclusion probability $[0.0, 1.0]$ from grey (non-active) to blue (active). The loss function encourages the network to minimize both the reconstruction error and the SINDy loss in $\zv$ and $\xv$. The Bayesian prior works as a regularization term on $\Xiv$ for sparse inference.}
\label{fig:overview}
\end{figure*}
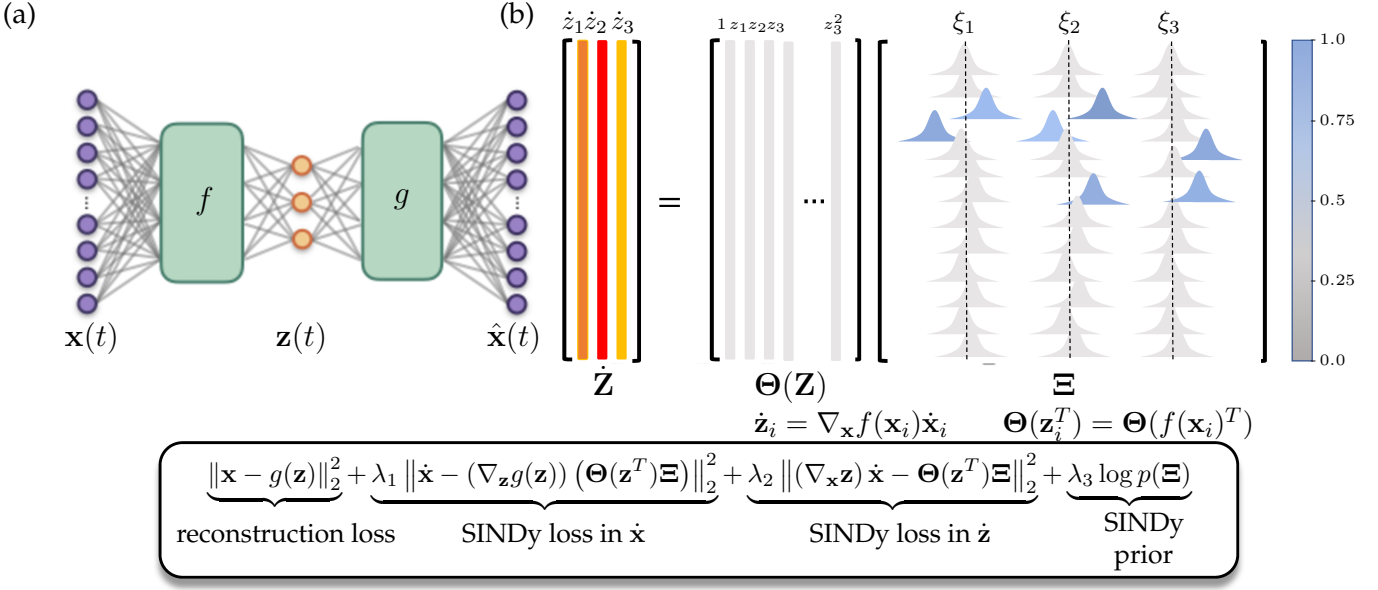

The SINDy autoencoder enables a joint discovery of sparse dynamical models and coordinates. Figure~\ref{fig:overview} (a) provides an overview of an autoencoder. The input data $x(t) \in \real^d$ is mapped by an encoder function $f_{\theta_1}(\cdot)$ to a latent space $z(t) \in\real^{d_z}, d_z < d$. This latent space $z(t)$ contains sufficient information to recover $x(t)$ via a decoder function $g_{\theta_2}(\cdot)$. 

SINDy autoencoder combines SINDy with autoencoders by constraining the latent space governed by a sparse dynamical system. The encoder function $f_{\theta_1}(\cdot)$ performs coordinate transformation to map the high-dimensional inputs into an appropriate latent subspace.   The latent space $z(t)=f_{\theta_1}(x(t))$ has an associated sparse dynamical model governed by 
\begin{align}
    \frac{d}{dt} z(t) = \Phi(z(t)) = \Thetav(z(t))\Xi, 
\end{align}
where $\Thetav(z)=[\theta_1(z), \theta_2(z),\dots,\theta_p(z)]$ is a library of candidate basis functions, and a set of coefficients $\Xi=[\xi_1, \xi_2, \dots, \xi_p]$. 

A statistical understanding of the model formulates the SINDy autoencoder as a parametric model $\mathcal{M}_\theta$ where $\theta=\{\theta_1, \theta_2, \Xi\}$ contains all parameters. The likelihood of this model $p(\mathcal{D}|\theta)$ is defined as
\begin{align}
    p(\mathcal{D}|\theta) \propto \exp\lrp{\left\| \xv - g_{\theta_2}(\zv)\right\|_2^2 + \lambda_1 \left\|\dot{\xv} - \left(\nabla_\zv g_{\theta_2}(\zv)\right)\left(\Thetav(\zv^T)\Xiv\right)\right\|_2^2 + \lambda_2 \left\|\left(\nabla_\xv\zv\right)\dot{\xv} - \Thetav(\zv^T)\Xiv\right\|_2^2 }. 
\end{align}
The log-likelihood of this statistical model is similar to the setting in~\citep{champion2019data}. 
In order to promote sparsity on $\Xi$, we consider two sets of priors in what follows. 

\paragraph{Laplace prior.} The Laplace prior can be understood as a Bayesian LASSO~\cite{tibshirani1996regression,park2008bayesian}. We define the Laplace prior such that
\begin{align}
    \label{eqn:laplace_def}
    \Xi_j \sim \mathcal{L}(0, v_0), 
\end{align}
where $\mathcal{L}(\cdot, \cdot)$ denotes the Laplace distribution defined as $f(\Xi_i;0,v_0) = \frac{1}{2v_0}\exp\lrp{-\frac{|\Xi_i|}{v_0}}$. We can see the equivalence of the Laplace prior and the LASSO in the negative log-likelihood, where $\frac{1}{v_0}||\Xi_i||_1$ is included as a regularizer. 

\paragraph{Spike-and-slab Gaussian-Laplace (SSGL) prior.} We define the SSGL prior as 
\begin{align}
    \label{eqn:ssgl_def}
    \Xi_j|\gamma_j \sim (1-\gamma_j) \mathcal{L}(0, \sigma v_0) + \gamma_j \mathcal{N} (0, \sigma^2 v_1), 
\end{align}
where $\gamma_j$ is a binary variable, $\Xi \in\real^p$, $\sigma, v_0, v_1 \in\real$, $\mathcal{L}(\cdot,\cdot)$ denotes a Laplace distribution and $\mathcal{N}(\cdot, \cdot)$ denotes a Normal distribution. We assign a Bernoulli prior to $\gamma\sim Ber(\delta)$, $\delta\in [0,1]$. The prior for $\theta_1,\theta_2$ is specified with a Gaussian, which is equivalent to an $\ell_2$ regularization implementation-wise. For simplicity, we set $\delta, \sigma, v_0, v_1$ as tunable hyperparameters.

\paragraph{Prior selection.}
In general, there is no optimal Bayesian prior for all statistical models since every prior has its own advantages and drawbacks. Therefore, the prior selection typically requires an assessment of both theory and experimental outcomes. If one selects the prior of $\Xi$ to be a Laplace distribution, the setting will be identical to the basic SINDy autoencoder model, which is equivalent to adding a $\ell_1$ regularization term (c.f. Eqn. (7) and Figure 1 in~\citep{champion2019data}). Even if the Laplace prior has a benefit in computation, its performance suffers for small sample sizes and large observation noises. Different from the Laplace prior, if one selects the SSGL prior for $\Xi$, it typically requires a slightly increased cost in computation. However, the SSGL prior typically has dominating performances for cases with very large noise and limited sample size, which is preferred in our case to learn the actual video data. 

\paragraph{Bayesian formulation.}
Using the Bayes formula, we can construct the posterior distribution from the likelihood function and prior that
\begin{align}
    \pi(\theta,\gamma|\mathcal{D}) \propto p(\mathcal{D}|\theta) p(\theta_1)p(\theta_2)p(\Xi|\gamma)p(\gamma). 
\end{align}
We aim to depict the posterior distribution $p(\Xi|\mathcal{D})$ via the joint distribution $p(\theta|\mathcal{D})$ under the sparsifying SSGL prior. The approximation of $p(\Xi|\mathcal{D})$ is accessible from posterior samples of $p(\theta|\mathcal{D})$ when dropping $\theta_1,\theta_2$. In the setting of deep neural networks, we can sample the posterior distribution via Stochastic Gradient methods using mini-batches. The mini-batch setting not only naturally fits into the training of deep neural networks but also could accelerate the Bayesian posterior sampling process.

\subsection{Stochastic Gradient Langevin Dynamics}  \label{sec:sgld}
To perform posterior sampling in mini-batch settings, Stochastic Gradient Langevin Dynamics (SGLD) is a  popular method that combines stochastic optimization and Langevin Dynamics~\citep{welling2011bayesian}. 
Denote the learning rate at epoch $t$ by $\epsilon^{(t)}$ which decreases to zero, and the dataset $\mathcal{D}=\{d_i\}_{i=1}^N$. The mini-batch setting estimates the gradient $\nabla_\theta L(\theta)$ from a subset (batch) $\mathcal{B} = \{d_i\}_{j=1}^{n}$. 
The injected noise from mini-batches facilitates the generation of posterior samples while reducing the high computational cost in computation for full-batch gradients. 
We follow the classical SGLD setting where
\begin{align}
    \Delta\theta_{t+1}=\frac{\epsilon^{(t)}}{2}\left(\nabla\log p(\theta_{t})+\frac{N}{n}\sum_{i=1}^{n}\nabla\log p(X_{i}|\theta_{t})\right)+\eta_{t},\;\;\;\;\eta_{t}\sim\mathcal{N}(0,\epsilon^{(t)}).
\end{align}
Here, $p(\theta_t)$ denotes the prior specified in Eqn.~\eqref{eqn:laplace_def},~\eqref{eqn:ssgl_def} and $p(X_i|\theta_t)$ denotes the data likelihood. Prior works study the asymptotic convergence of SGLD to the target distribution, which validates SGLD for posterior sampling in theory~\citep{zhang2017hitting,teh2016consistency,hoffman2020black}. An advantageous property of SGLD in posterior sampling is that with decaying step size $\epsilon^{(t)}$, SGLD automatically transfers from a stochastic optimization algorithm to a posterior sampling procedure (c.f. Sec. 4.1~\citep{welling2011bayesian}). The Metropolis-Hasting correction can be ignored since the rejection rate for sampling goes to zero asymptotically, resulting from $\epsilon^{(t)} \to 0$ when $t\to\infty$. The discretization error similarly decreases as $\epsilon$ goes to zero. 

\paragraph{Cyclical SGLD.} The cyclical SGLD method~\citep{zhang2019cyclical} consists of exploration and sampling stages via a cyclical step-size schedule for $\epsilon^{(t)}$. In the training of deep autoencoders, the optimization process is highly non-convex with a very complex loss landscape. The cyclical step-size schedule helps to explore the parameter space when $\epsilon$ is large as well as sample local mode when $\epsilon$ is small. It is also possible to understand cyclical SGLD from a parallel SGLD perspective~\citep{deng2020non}. We apply this idea in our experiments to perform better inference. 

\subsection{Empirical Bayes Variable Selection in SINDy Autoencoder}

The empirical bayesian method infers prior hyperparameters from data. In this case, we aim to optimize $\gamma$ (ignoring the uncertainty) and sample $\theta|\mathcal{D}$. The posterior distribution of $\Xi | \mathcal{D}$ can be derived from posterior samples of $\theta|\mathcal{D}$. 

Using the mini-batch setting, the posterior distribution follows
\begin{align}
    \pi(\theta,\gamma|\mathcal{B}) \propto p(\mathcal{B}|\theta)^{\frac{N}{n}} p(\theta_1)p(\theta_2)p(\Xi|\gamma)p(\gamma). 
\end{align}
The term $p(\mathcal{B}|\theta)$ can be evaluated from the loss of SINDy autnencoder for the current mini-batch $\mathcal{B}$; the term $p(\theta_1),p(\theta_2)$ can be computed from the $\ell_2$ loss; the term $p(\Xi|\gamma)$ can be computed from Eqn.~\eqref{eqn:ssgl_def}; and the term $p(\gamma)$ can be known from the Bernoulli prior setting.

As a binary variable, $\gamma$ is difficult to be optimized for due to non-continuity and non-convexity. An important trick to perform the optimization on $\gamma$ is to alternatively optimize the adaptive posterior mean $\Ep{\gamma|\theta^{(k)},\mathcal{D}}{\pi(\theta,\gamma|\mathcal{D})}$ which treats $\gamma$ as a latent variable. Following previous works with similar settings~\citep{deng2020non,rovckova2014emvs}, we could indirectly evaluate $\pi(\theta,\gamma|\mathcal{D})$ by a strict lower bound $Q(\cdot\mid\cdot)$ that
\begin{align}
    \label{eqn:q_def}
    Q(\theta|\theta^{(k)}) &= \Ep{\mathcal{B}}{\Ep{\gamma\mid\theta^{(k)}\mathcal{D}}{\log\pi(\theta,\gamma|\mathcal{B})}} \leq \log\Ep{\gamma|\theta^{(k)},\mathcal{D}}{\Ep{\mathcal{B}}{\pi(\theta,\gamma|\mathcal{B})}}. 
\end{align}
The inequality holds by Fubini’s theorem and Jensen’s inequality (c.f. Eqn. (7) in~\citep{deng2020non}). 

The variable $Q(\theta|\theta^{(k)})$ can be decomposed into 
%
\begin{align}
    Q(\theta|\theta^{(k)}) &= \frac{N}{n}\log\pi(\mathcal{B}|\theta) - \lrn{\theta_1} - \lrn{\theta_2} -\sum_{i\in|\Xi|}\Biggl[
    \frac{|\Xi_{i}|\kappa_{i0}}{\sigma}+\frac{\Xi_{i}^{2}\kappa_{i1}}{2\sigma^{2}}
    \Biggl] + \sum_{i\in |\Xi|}\log\lrp{\frac{\delta}{1-\delta}}\rho_{i}+C,
\end{align}
where $\kappa_{i0} = \mathbb{E}_{\gamma|\theta^{(k)},\mathcal{D}}\lrb{\frac{1}{v_{0}(1-\gamma_{i})}}$, $\kappa_{i1} = \mathbb{E}_{\gamma|\theta^{(k)},\mathcal{D}}\lrb{\frac{1}{\gamma_{i}}}$, $\rho_i = \Ep{\gamma|\theta^{(k)},\mathcal{D}}{\gamma_i}$, and $C\in\real$ is a constant. 
Notice here $\gamma$ is treated as a latent variable, and we only consider the expectation given the conditional distribution of $\gamma$ given $\theta^{(k)},\mathcal{D}$. The $\rho_i$ could be considered as an inclusion probability estimate of $\Xi_i$. 
In this way, $\rho$ softens $\gamma$ from a binary variable into a continuous one. 

The term $\kappa$ performs an elastic net-like approach which adaptively optimizes the $\ell_1$ and $\ell_2$ coefficients. Suppose $\Xi_i$ is identified with a high probability to be a sparse variable (e.g., $\rho_i<0.05$), the term $\kappa_{i0}$ will be large, strengthening the sparsity constraints. Otherwise, if $\Xi_i$ is identified as a non-sparse variable (e.g. $\rho_i>0.95$), the term $\kappa_{i1}$ will be small, and the $\ell_2$ constraint will be dominated instead.

\subsubsection{Stochastic Approximation from Expectation Maximization of \texorpdfstring{$\rho$}{rho} and \texorpdfstring{$\kappa$}{kappa}}

We could derive an asymptotically correct posterior distribution on $\pi(\theta, \kappa, \rho)$ following the steps in~\citep{deng2020non}: 
\begin{enumerate}
    \item Sample $\theta$ from $Q(\cdot)$ that 
    \begin{align}
    \theta^{(k+1)} = \theta^{(k)} + \eta^{(k)}\nabla_\theta Q(\rho^{(k)}, \kappa^{(k)} | \mathcal{B}^{(k)}) + \mathcal{N}(0, 2\eta^{(k)})
    \end{align}
    \item Perform stochastic approximation to latent variables $\rho,\kappa$ that
    \begin{align}
        \rho^{(k+1)} &= (1-\omega^{(k+1)})\rho^{(k)} + \omega^{(k+1)}\tilde{\rho}^{(k+1)},\\
        \kappa^{(k+1)} &= (1-\omega^{(k+1)})\kappa^{(k)} + \omega^{(k+1)}\tilde{\kappa}^{(k+1)}.
    \end{align}
\end{enumerate}
Here, $\tilde{\rho}^{(k+1)},\tilde{\kappa}^{(k+1)}$ is the Expectation Maximization (EM) estimation of $\rho,\kappa$ given iteration $k$~\citep{rovckova2014emvs}. The EM estimation of the inclusion probability $\rho_i$ is
\begin{align}
    \tilde{\rho}^{(k+1)}_{i}=\mathbb{E}_{\gamma|\theta^{(k)},\mathcal{B}}[\gamma_{i}]=P(\gamma_{i}=1|\Xi^{(k)})=\frac{a_{i}}{a_{i}+b_{i}},
\end{align}
where $a_{i}=\pi(\Xi_{i}^{(k)}|\gamma_{i}=1)P(\gamma_{i}=1|\delta)$
and $b_{i}=\pi(\Xi_{i}^{(k)}|\gamma_{i}=0)P(\gamma_{i}=0|\delta)$. In our case, the term $\pi(\Xi_i^{(k)}|\gamma_i=1)$ is the probability of $\Xi_i^{(k)}$ given the Gaussian prior distribution, and the term $\pi(\Xi_i^{(k)}|\gamma_i=0)$ is the probability of $\Xi_i^{(k)}$ from the Laplace distribution. The latter term $P(\gamma_{i}=1|\delta)=\delta$ given Bernoulli prior in the previous setting.  
Following a similar process, for $\kappa$, we have
\begin{align}
    \tilde{\kappa}_{i0}^{(k+1)} &= \mathbb{E}_{\gamma|\theta^{(k)},\mathcal{B}}\lrb{\frac{1}{v_{0}(1-\gamma_{i})}}=\frac{1-\rho_{i}}{v_{0}}, \\
    \tilde{\kappa}_{i1}^{(k+1)} &= \mathbb{E}_{\gamma|\theta^{(k)},\mathcal{B}}\lrb{\frac{1}{v_1\gamma_{i}}}=\frac{\rho_{i}}{v_1}.
\end{align}

\subsection{Prediction with Bayesian SINDy Autoencoder}

\begin{figure*}
\vspace{-.1in}
\includegraphics[width=\textwidth]{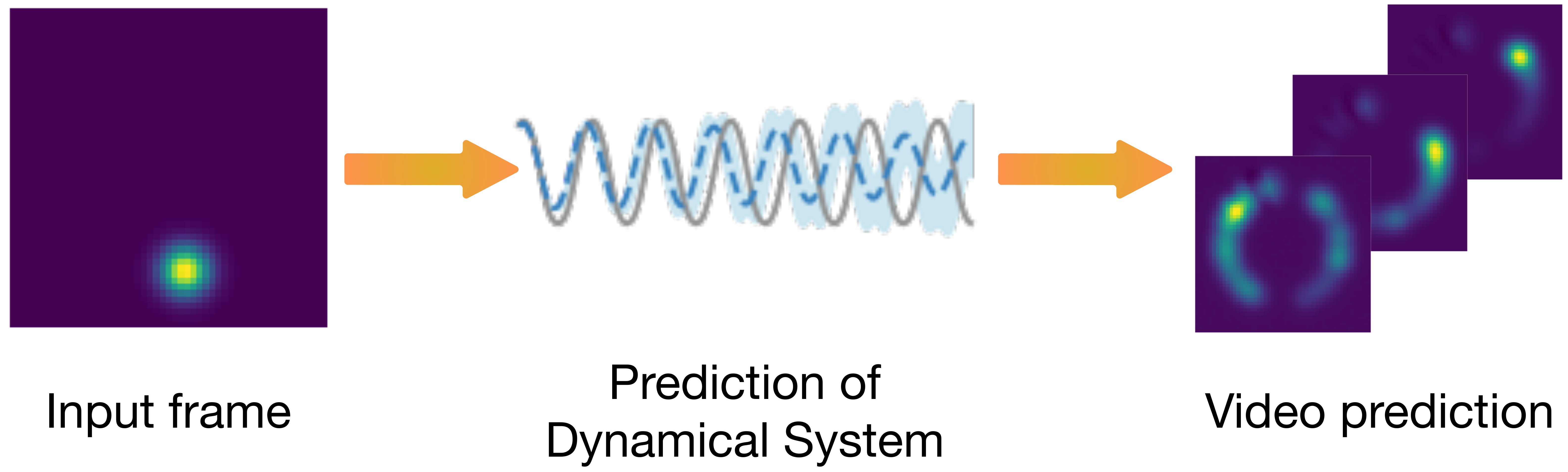}
 \centering
\caption{Add more things to this figure including: encoder (with its shape), decoder (with its shape), and Bayesian SINDy.}
\label{fig:video_pred}
\end{figure*}

Bayesian SINDy autoencoder has a trustworthy application in our application of video prediction whereby we learn the dynamics and coordinates of the latent spae. Based on an inference process established in previous sections, the video prediction can precisely understand the underlying dynamical system, which allows accurate, robust, and interpretable future forecasting. 
Additionally, the Bayesian framework enables precise uncertainty quantification for video prediction from posterior samples. We visualize the prediction process in Fig.~\ref{fig:video_pred}. In this case of a pendulum video, the uncertainty of video prediction grows with larger $t$, and gradually fails to predict by showing a ring-like prediction. 

From the procedure in Sec. 3.3, we can generate samples from $p(\theta|\mathcal{D})$, which is the posterior distribution of neural network parameters $\theta$ given observed data $\mathcal{D}$. Suppose the posterior samples are $\Xi^1,\Xi^2,...,\Xi^m \sim p(\theta|\mathcal{D})$. The full posterior predictive distribution is defined as
\begin{align}
    p(\hat{x}(t)|x_0,\mathcal{D}) &= \int p(\hat{x}(t)|\theta, x_0) p(\theta|\mathcal{D}) d\theta\nonumber\\
    &= \int p(\hat{x}(t)|\theta_2, \hat{z}(t)) 
    p(\hat{z}(t)|\Xi,\hat{z}(0))
    p(\hat{z}(0)|\theta_1,x_0)p(\theta_1,\theta_2,\Xi|\mathcal{D}) d\theta
\end{align}

Therefore, we can approximately generate samples from $p(\hat{x}(t)|x_0,\mathcal{D})$ using Monte Carlo estimation. For simplicity, we only consider the Maximum likelihood Estimation $\theta_1^{\text{MLE}},\theta_2^{\text{MLE}}$. We have the following process:  
\begin{enumerate}
    \item From the input image $x_0$, compute $z(0)=f_{\theta_1^{\text{MLE}}}(x_0)$. 
    \item From $z(0)$, using posterior samples of $\Xi$, generate $\hat{z}^{(i)}(t)$ samples via
    \[
    z^{(i)}(t) = z(0) + \int_0^t \Theta(z(t'))\Xi^{(i)}dt'.
    \]
    \item From samples of $z^{(i)}(t)$, generate $\hat{x}^{(i)}(t)=g_{\theta_2^{\text{MLE}}}(z^{(i)}(t))$. 
\end{enumerate}

\section{Experiments}

In the following subsections, we conduct four case studies on Bayesian SINDy autoencoder for governing equations and coordinate system discovery for video data. 
In Sec.~\ref{sec:expr_synthetic}, we study similar cases to those in~\citep{champion2019data} with synthetic high-dimensional data generated from the Lorenz system, reaction-diffusion, and a single pendulum. We explore the Laplace prior for chaotic Lorenz system and the SSGL prior for reaction-diffusion and single pendulum. 
In Sec.~\ref{sec:expr_real}, we study real video data that consists of 390 video frames of a moving rod. This setting is particularly challenging due to the high dimensionality in video data, noisy observation, missing temporal derivatives, and prior setting for correct learning. 
In both experiments on synthetic and real video data, we observe the Bayesian SINDy autoencoder can accurately perform nonlinear identification of dynamical system under the correct setting of preprocessing, prior, and training parameters.
All the experiments are implemented and run on a single NVIDIA GeForce RTX 2080 Ti.

\subsection{Learning Physics from Synthetic Video Data}
\label{sec:expr_synthetic}
\subsubsection{Chaotic Lorenz System via the Laplace prior}
We consider the chaotic Lorenz system in the following
\begin{align}
\begin{cases}
\dot{z}_1 = -\sigma z_1 + \sigma z_2 \\
\dot{z}_2 = \rho z_1 - z_2 - z_1z_3\\
\dot{z}_3 = z_1z_2 -  \beta z_3,
\end{cases}    
\end{align}
where $z=[z_1,z_2,z_3]\in\real^3$ and $\sigma,\rho,\beta$ are constants. 
The Lorenz system is very representative of the chaotic and nonlinear system, which is an ideal example of applying model discovery techniques. 
In the numerical simulation, we first set $\sigma=10, \rho=28, \beta=-2.7$. We only generate partial Lorenz via time range $t=[0, 5]$ with $\Delta t = 0.02$ for $1024$ different Lorenz systems from random initial conditions. The initial condition follows a uniform distribution centered at $[0,0,25]$ with width $[36, 48, 41]$ respectively. 

From the underlying dynamical system, we create a high-dimensional dataset via six fixed spatial models given by Legendre polynomials that $\ve{u}_1, \ve{u}_2, ..., \ve{u}_6\in\real^{128}$. 
We transfer from the low-dimensional dynamical system into a high-dimensional dataset via the following rule: 
\begin{align}
    \ve{x}(t) = \ve{u}_1 z_1(t)+ \ve{u}_2 z_2(t)+ \ve{u}_3 z_3(t)+ \ve{u}_4 z_1(t)^3 + \ve{u}_5 z_2(t)^2 + \ve{u}_6 z_3(t)^3.
\end{align}
\begin{figure}
    \centering
    \includegraphics[width=\textwidth]{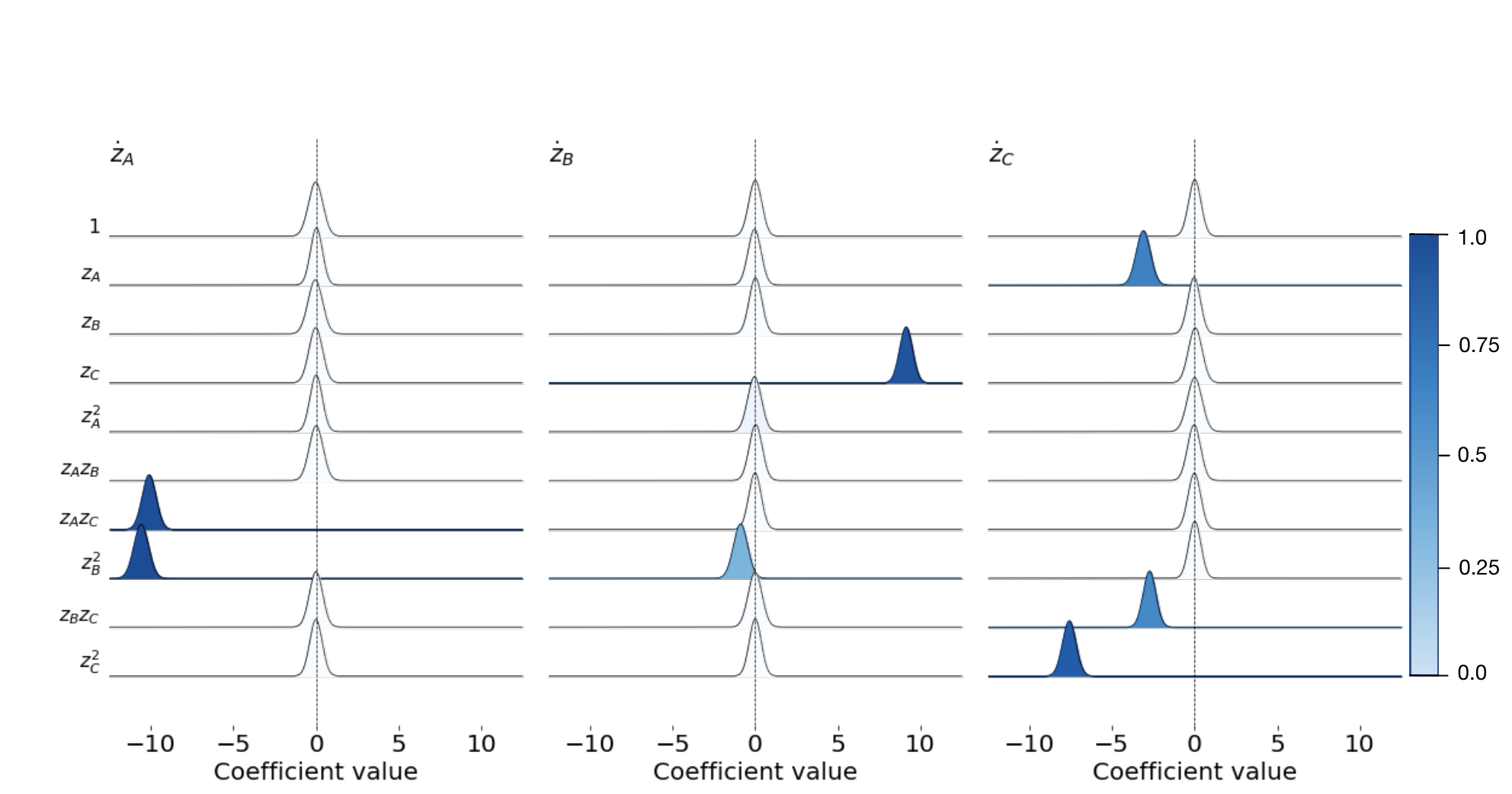}
    \caption{Bayesian estimation and uncertainty quantification visualization of SINDy coefficient for Lorenz system under Laplace prior. The color bar represents the inclusion probability estimate given the coefficient magnitude.  }
    \label{fig:lorenz_laplace}
\end{figure}
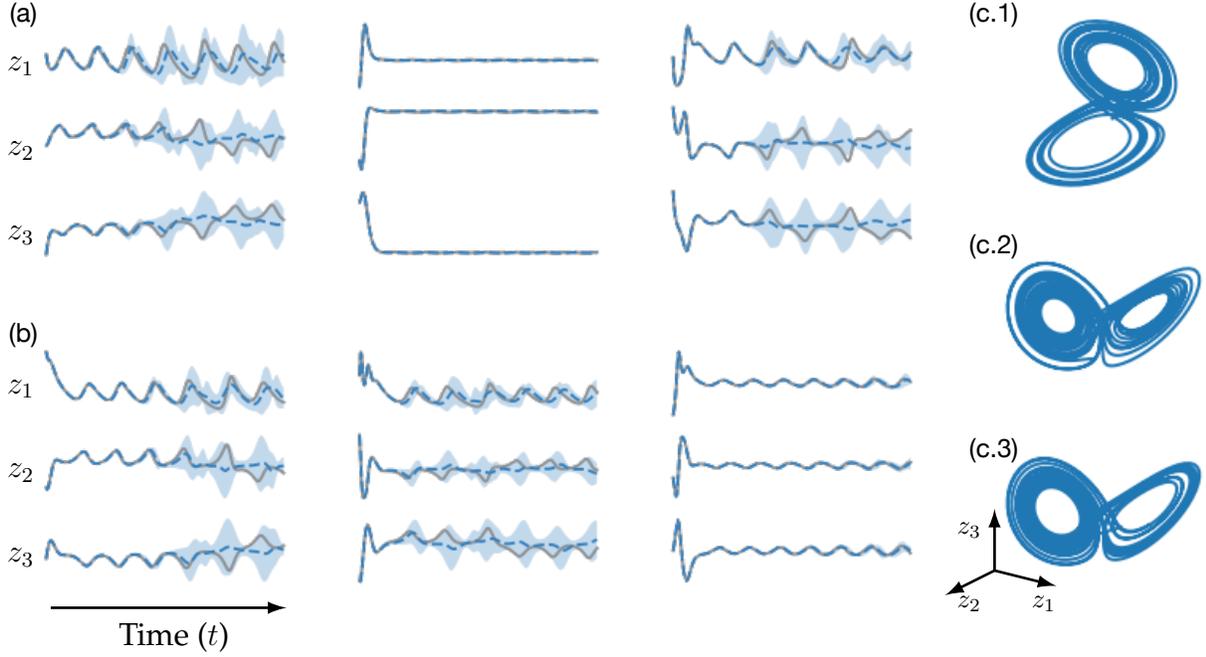
\begin{figure}
    \centering
    \begin{overpic}[width=\textwidth]{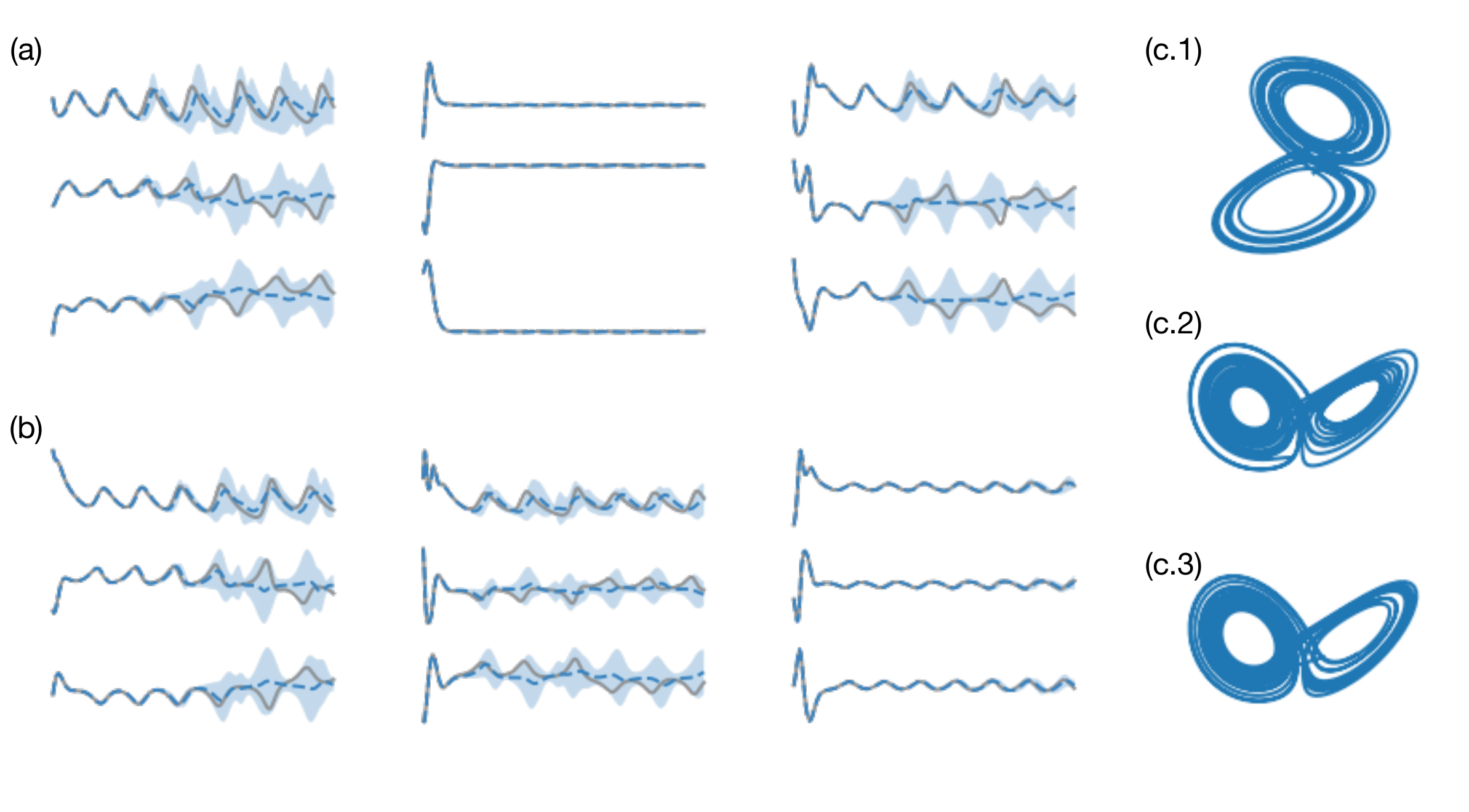}
    \linethickness{1pt}
    \put(0.5,21.3){\large $z_1$}
    \put(0.5,14.5){\large $z_2$}
    \put(0.5,7.7){\large $z_3$}
    \put(4.0,4){{\color{black}\vector(1,0){19}}}
    \put(9.5,1){\large Time ($t$)}
    \put(0.5,47.3){\large $z_1$}
    \put(0.5,40.5){\large $z_2$}
    \put(0.5,33.7){\large $z_3$}
    \put(80,7){{\color{black}\vector(0,1){5}}}
    \put(80,7){{\color{black}\vector(1,-0.25){5}}}
    \put(80,7){{\color{black}\vector(-1,-0.5){4}}}
    \put(77,10){\small $z_3$}
    \put(77,4){\small $z_2$}
    \put(83,4){\small $z_1$}
    \end{overpic}
    \caption{(a) Trajectory prediction for \textbf{in-distribution} data with uncertainty quantification generated by posterior samples from Bayesian inference on SINDy coefficient. (b) Trajectory prediction for \textbf{out-of-distribution} data with uncertainty quantification generated by posterior samples from Bayesian inference on SINDy coefficient. (c.1) Discovered Lorenz systems and (c.2) transformation to the standard space. (c.3) Lorenz system generated from the ground truth. }
    \label{fig:lorenz_prediction}
\end{figure}
We set the autoencoder with latent dimension $d=3$ corresponding to the latent system in the 3D coordinate system with $z_A, z_B, z_C$. We include polynomials with the highest order $3$ composing a library $[1, z_A, z_B, z_C, z_A^2, z_Az_B, z_Az_C, z_B^2, z_Bz_C, z_C^2, z_A^3, z_A^2z_B, z_A^2z_C, ..., z_C^3]$. 
Via the autoencoder, we wish to identify the correct active terms as well as the value of the coefficients. 
The coefficient of $\Xi$ is uniformly initialized from constant $1$. The loss coefficients are $\lambda_1=0.0$, $\lambda_2=1\times 10^{-4}$. For the encoder and decoder, we use the sigmoid activation function with widths $[64, 32]$. For optimization, we select Adam optimizer with learning rate $1\times 10^{-3}$ and the batch size to be 1024. For the Laplace prior setting, we set $\lambda_3=1\times 10^{-5}$. 

By training with $5,000$ epochs following the setting from~\citep{champion2019data}, in terms of the error metrics, the best test error of the decoder reconstruction achieves $2\times 10^{-5}$ of the fraction of the variance from the input. 
The fraction of unexplained variances are $2\times 10^{-4}$ for the reconstruction of $\dot{z}$, and $1.3\times 10^{-3}$ for the reconstruction of $\dot{x}$. 
Notice here decoder reconstruction is better compared to SINDy autoencoder without uncertainty quantification, but the reconstruction of $\dot{z}$ is slightly worse compared to point estimation for SINDy autoencoder. 

The coefficient estimate from the Bayesian SINDy autoencoder is shown in Fig.~\ref{fig:lorenz_laplace}. From the figure, we can observe the uncertainty quantification in the parameter space. The outcome of this model identifies $7$ active terms marked with deeper blues. 
It is known that the identification of Lorenz dynamics suffers from the symmetry in the coordinate system as described in~\citep{champion2019data}. The discovered governing equation could be equivalently transformed via the affine group transformation on the coefficients and a permutation group transformation on the latent variables. Therefore, the identification is still correct in Fig.~\ref{fig:lorenz_laplace} from the following transformations. (a) We inversely transform the permutation group via assigning $z_C$ to $z_1$, $z_B$ to $z_2$, and $z_A$ to $z_3$. (b) We inversely perform affine transformation by $z_1=1.0$, $z_2=-0.94z_2$, and $z_3=0.55z_3-2.81$. We demonstrate the effectiveness of the affine transformation process in Fig.~\ref{fig:lorenz_prediction} (c) that the discovered model (c.1) can be transformed into (c.2), which is close to the ground truth model (c.3). The discovered governing equation from the Laplace prior is 
\begin{align}
    \begin{cases}
    \dot{z}_1 = -10.09 z_1 + 10.00 z_2 \\
    \dot{z}_2 = 27.09 z_1 - 0.86 z_2 - 5.35z_1z_3\\
    \dot{z}_3 = 5.35z_1z_2 -  2.71 z_3,
    \end{cases}   
\end{align}

Stepping further from the uncertainty quantification of coefficients, we can see the uncertainty quantification in the prediction space in Fig.~\ref{fig:lorenz_prediction} (a) (b). In general, the uncertainty coverage correctly captures the true trajectory when the model fails to predict correctly and mildly covers the trajectory when the model prediction is confident. 
 
\subsubsection{Reaction Diffusion via Spike-and-slab prior}
Reaction-diffusion is governed by a partial differential equation (PDE) that has complex interactions between spatial and temporal dynamics. 
We define a lambda-omega reaction-diffusion system by
\begin{align}
    u_t = (1-(u^2+v^2))u + \beta(u^2+v^2)v + d_1(u_{xx}+u_{yy}) \nonumber\\
    v_t = -\beta(u^2+v^2)u + (1-(u^2+v^2))v + d_2(v_{xx}+v_{yy}),
    \label{eqn:rd_synthetic}
\end{align}
which $d_1=d_2=0.1, \beta=1$. 

For the synthetic data, we first generates the latent dimensions $u(x,y,t)$ and $v(x,y,t)$ by~\ref{eqn:rd_synthetic} from $t=[0, 500]$ with time step $\Delta t=0.05$. Then, we generate the snapshots of the dynamical system spatially in $xy$-domain from its latent dimensions into a video with shape $(10000, 100, 100)$. We obtain a training dataset with size $9,000$. We also generate a testing dataset with $1,000$ samples. 

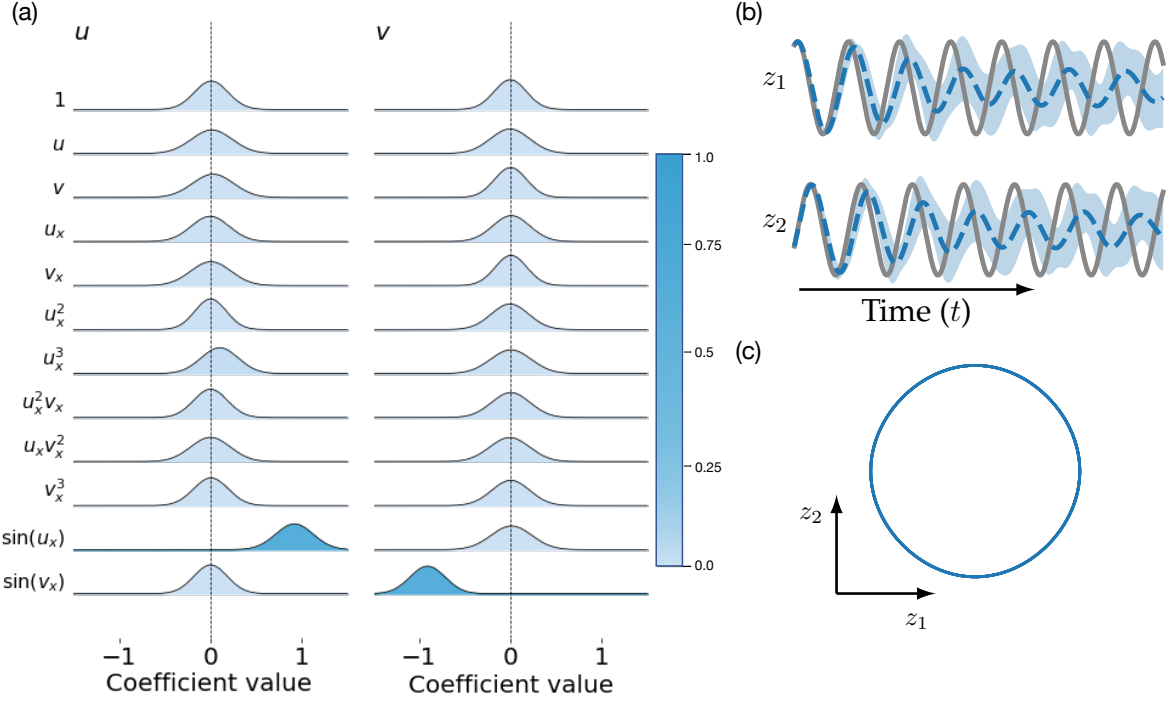
\begin{figure}
    \centering
    \begin{overpic}[width=\textwidth]{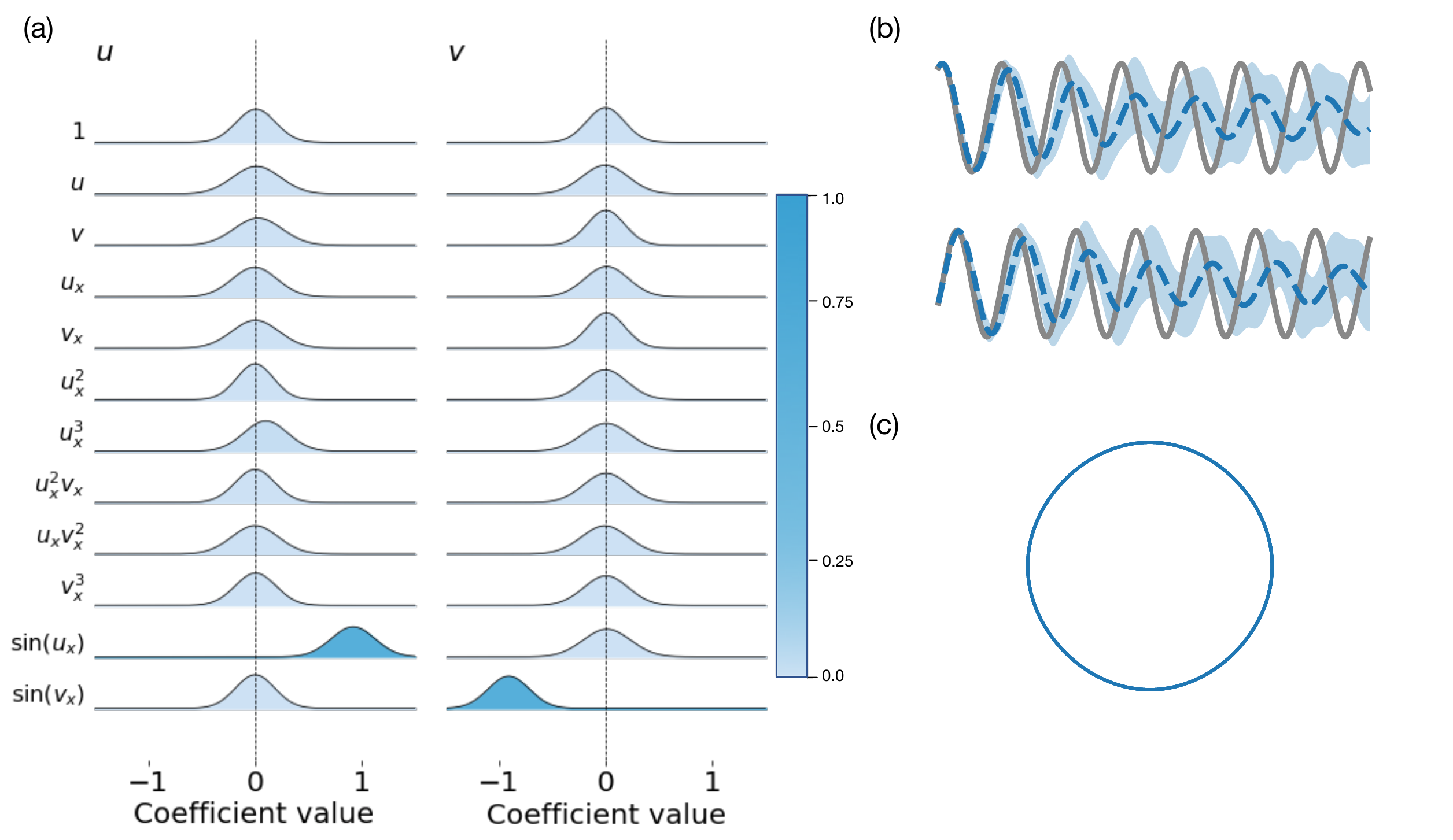}
    \linethickness{1pt}
    \put(68,9){{\color{black}\vector(0,1){8}}}
    \put(68,9){{\color{black}\vector(1,0){8}}}
    \put(65,33.5){{\color{black}\vector(1,0){19}}}
    \put(70,31){\large Time ($t$)}
    \put(65,15){\small $z_2$}
    \put(73.5,6.5){\small $z_1$}
    \put(62,50.0){\normalsize $z_1$}
    \put(62,38.5){\normalsize $z_2$}
    \end{overpic}
    \caption{(a) Bayesian estimation and uncertainty quantification visualization of SINDy coefficient for reaction-diffusion under Spike-and-slab prior. (b) Visualization of in-distribution (top) and out-of-distribution (bottom) predicted dynamics. (c) The generated attractor from the mean of Bayesian inference. }
    \label{fig:rd_bayes}
\end{figure}

We setup the autoencoder with latent dimension $d=2$, targeting the $u, v$, and apply a first-order library of functions including $[1, u_x, v_x, u_x^2, u_xv_x, v_x^2, u_x^3, u_x^2v_x, u_xv_x^2, v_x^3, \sin(u_x), \sin(v_x)]$. We hope to discover the two oscillating spatial modes for this nonlinear oscillatory dynamics. 
The coefficient of $\Xi$ is randomly initialized from Gaussian $\mathcal{N}(0, 0.1)$. The loss coefficients are $\lambda_1=1.0\times 10^{-2}, \lambda_2=1.0\times 10^{-1}, \lambda_3=20.0$. 
For the encoder and decoder, we use sigmoid activation function with width $[256]$. For optimization, we select Adam optimizer with learning rate $1e^{-3}$, and batch size to be $1000$.  For the setting of SSGL prior, we set $\delta=0.08$, $v_0=0.1$, $v_1=3.0$, $\omega^{(k)}=0.02\times (0.999)^k$. 

We follow the same setting in~\citep{champion2019data}. The SSGL prior only requires $1,500$ epochs for training while LASSO setting frequently needs more than $3,000$ epochs for training. The best test error is $2.1\times 10^{-3}$ for decoder loss, $2.1\times 10^{-5}$ for the reconstruction of $\dot{z}$, and $1.7\times 10^{-4}$ for the reconstruction of $\dot{x}$. 
The fraction of unexplained variance of decoder reconstruction is $9.1\times 10^{-5}$. 
For SINDy predictions, the fraction of unexplained variance of $\dot{x}$ is $0.013$. The fraction of unexplained variance of $\dot{z}$ is $0.001$. These results all improve from LASSO based SINDy Autoencoder~\citep{champion2019data}. 
The posterior samples from Bayesian SINDy are shown in Fig.~\ref{fig:rd_bayes}. 
The discovered governing equation from the Bayesian SINDy autoencoder is 
\begin{align}
    \begin{cases}
    \dot{u}_x = 0.91\sin(u_x) \\
    \dot{z}_2 = -0.91\sin(v_x)
    \end{cases}   
\end{align}

\subsubsection{Nonlinear Pendulum via the Spike-and-slab prior}
We consider simulated video of a nonlinear pendulum in pixel space with two spatial dimensions. The nonlinear pendulum is governed by the following second-order differential equation:
\begin{equation} \label{eq:pendulum}
  \ddot{z} = -\sin z.
\end{equation}

We generate the synthetic dataset following the settings in~\citep{champion2019data}. The synthetic data first generates the latent dimension $z$ as the angle of pendulum from $t=[0, 10]$ with time step $\Delta t=0.02$. Then, we form a Gaussian ball around the mass point with angle $z$ and length $l$. This process transfers the dynamical system from its latent dimension $z$ into a video with shape $(500, 51, 51)$. By simulating this process with $100$ times, we obtain a training dataset with size $50,000$. We also generate a testing dataset with $5,000$ samples. 

We set up the autoencoder with latent dimension $d=1$, targeting the angle of pendulum, and apply a second-order library of functions including $[1, z, \dot{z}, z^2, z\dot{z}, \dot{z}^2, z^3, z^2\dot{z}, z\dot{z}^2, \dot{z}^3, \sin(z), \sin(\dot{z})]$. We hope to infer the real coefficient that $\Xi=[0, 0, 0, 0, 0, 0, 0, 0, 0, 0, -1, 0]$. The coefficient of $\Xi$ is randomly initialized from Gaussian. The loss coefficients are $\lambda_1=5e^{-3}, \lambda_2=5e^{-5}, \lambda_3=8e^{-4}$. For the encoder and decoder, we use the sigmoid activation function with widths $[128, 64, 32]$. For optimization, we select Adam optimizer with a learning rate $1e^{-3}$, and batch size to be $1000$.  For the setting of SSGL prior, we set $\delta=0.08$, $v_0=0.05$, $v_1=3.0$, $\omega^{(k)}=0.05\times (0.995)^k$. 

\begin{figure}
    \centering
    \begin{overpic}[width=\textwidth]{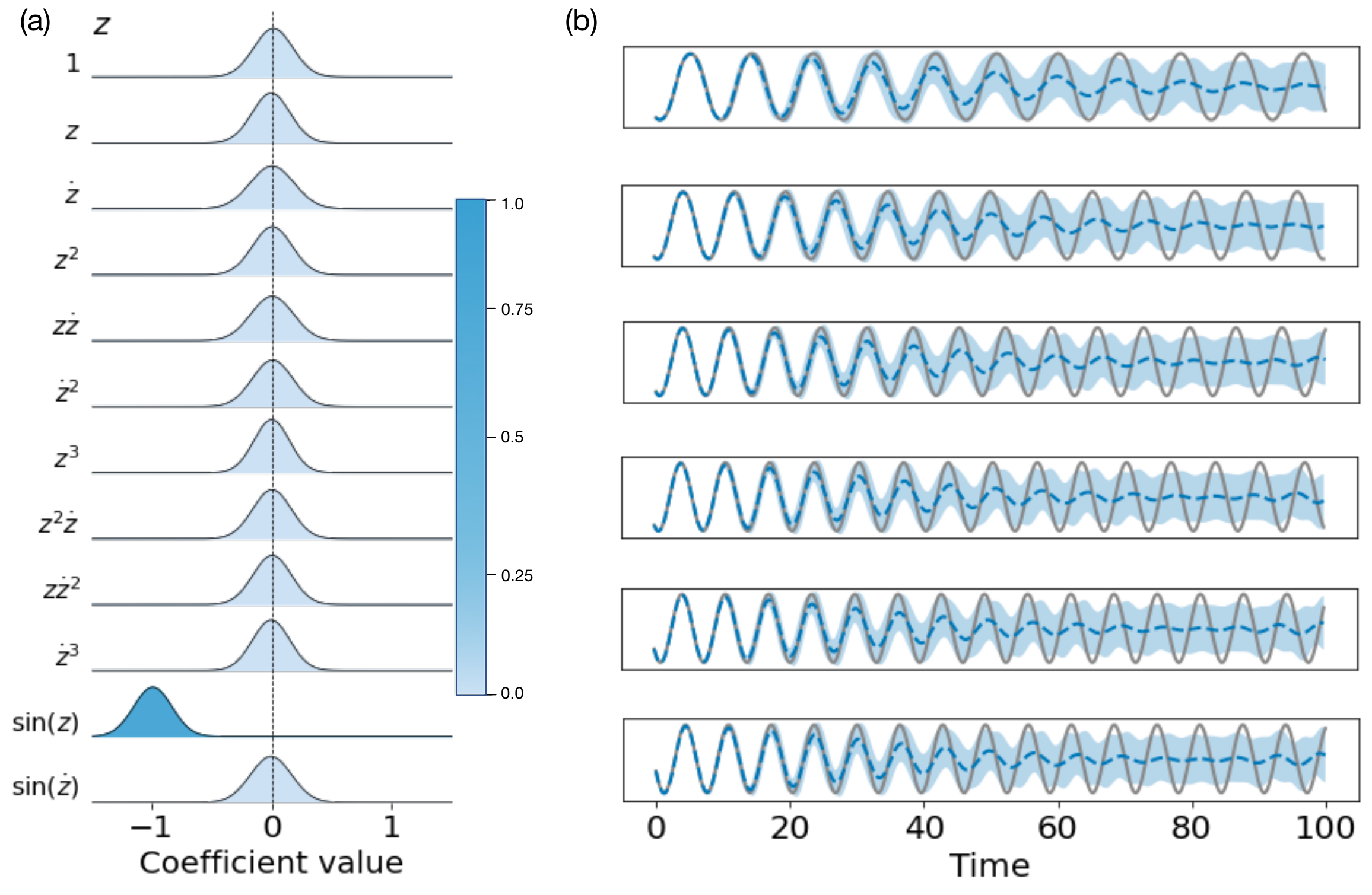}
    \put(41,9.0){\normalsize $z(t)$}
    \end{overpic}
    \caption{(a) Bayesian estimation and uncertainty quantification visualization of SINDy coefficient for pendulum data under Spike-and-slab prior. (b) Visualization of predicted dynamics with uncertainty quantification from 6 different initializations. }
    \label{fig:pen_bayes}
\end{figure}

We follow the same setting in~\citep{champion2019data}. The success discovery rate is $80\%$ from 15 training instances, which improves from $50\%$ success discovery rate in the LASSO setting. Additionally, Bayesian SINDy Autoencoder with the SSGL prior requires only $1,500$ epochs for training, while the LASSO setting frequently needs $5,000$ epochs for training. 
The best test error is $6.5\times 10^{-8}$ for decoder loss, $1.1\times 10^{-3}$ for the reconstruction of $\ddot{x}$, and $8.1\times 10^{-4}$ for the reconstruction of $\ddot{z}$. 
The best fraction of unexplained variance of decoder reconstruction is $4.5\times 10^{-4}$. 
For SINDy predictions, the best fraction of unexplained variance of $\ddot{x}$ and $\ddot{z}$ reconstruction are $2.3\times 10^{-4}$ and $5.5\times 10^{-3}$. The posterior samples from Bayesian SINDy are shown in Fig.~\ref{fig:pen_bayes} (a).
The discovered governing equation from the Bayesian SINDy autoencoder is 
\begin{align}
    \ddot{z} = -0.99\sin(z). 
\end{align}
The prediction of trajectory with uncertainty quantification is shown in Fig.~\ref{fig:pen_bayes} (b) with $6$ different initial conditions that $\theta_0=[\frac{2}{3}\pi, \frac{1}{2}\pi, \frac{1}{3}\pi, \frac{1}{4}\pi, \frac{1}{8}\pi, \frac{1}{16}\pi]$ from time $t=[0, 100]$. The error and uncertainty grow with longer time interval, generally starting from $t=20$. 

\subsection{Learning dynamical system from real video data}
\label{sec:expr_real}


\paragraph{Experimental setting and dataset description.} 
The raw video data has 14 seconds recording of a moving rod. 
We process the raw video via the following steps. 
\begin{figure}
    \centering
    \includegraphics[width=\textwidth]{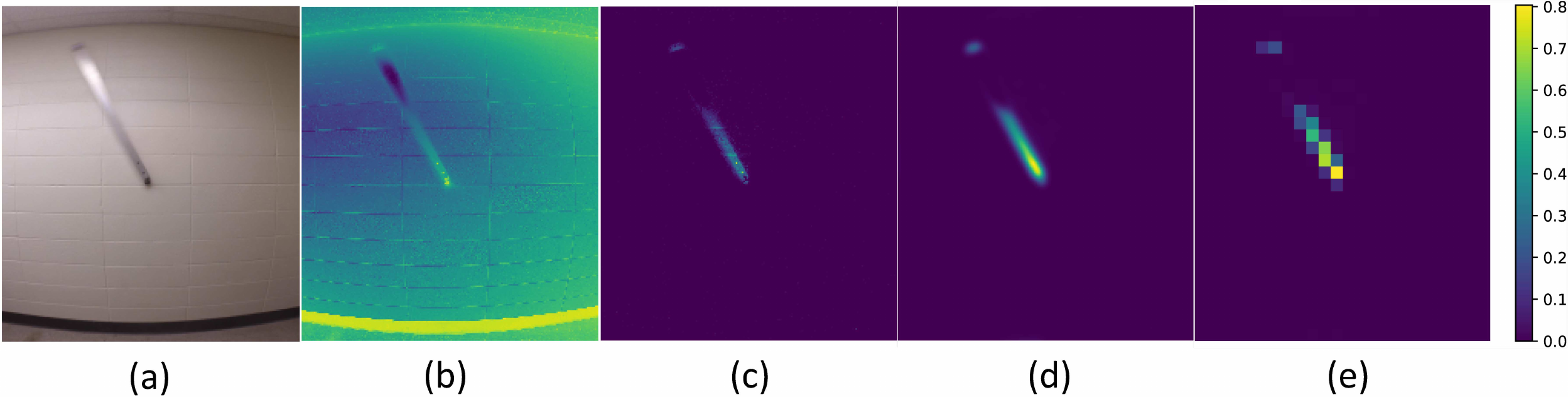}
    \caption{Preprocessing pipeline to high-dimensional video data to remove auxiliary information and reduce the size of video frames. }
    \label{fig:auxiliary_info}
\end{figure}
\begin{itemize}
    \item[(1)] We first move the RGB channels of a video to greyscale, normalizing to $[0, 1]$. This is step (a) to (b) in Fig. \ref{fig:auxiliary_info}. 
    \item[(2)] We estimate the background in this video. In our setting, we compute an averaged frame over the entire video as the background. In a wider context, one shall remove low-rank information via linear models \cite{li2004statistical,wright2022high}. This is step (b) to (c) in Fig. \ref{fig:auxiliary_info}. 
    \item[(3)] Apply Gaussian filters with appropriate variance. The selection of variance varies from case to case. A good hyperparameter setting should keep the mix-max gap of the original image and could also smoothen the sharp edges of the objects. This is step (c) to (d) in Fig. \ref{fig:auxiliary_info}. 
    \item[(4)] The final step downsample the frames from $(1080, 960)$ to $(27,24)$ via interpolation.  
\end{itemize}

After these preprocessing steps, we obtain a training dataset with 390 samples and shape $(27,24)$. 

\paragraph{Bayesian masked autoencoder setup}
We set up the autoencoder with latent dimension $d=1$, with a second-order library of functions similar to synthetic video data. We remove the constant term for simplicity of the inferential process. The loss coefficients are $\lambda_1=5e^{-7}, \lambda_2=5e^{-8}$, and $\lambda_3=1\times 10^{-4}$. For the encoder and decoder, we use the sigmoid activation function with widths $[64, 32, 16]$. For optimization, we select Adam optimizer with a learning rate $1e^{-3}$, and batch size to be $10$.  
Real video data is much harder compared to synthetic video data since it contains rich information, including object colors, background, and processing noises. At the same time, only $390$ snapshot samples are available to train the Bayesian autoencoder. To resolve this problem, we involve masked autoencoder~\citep{he2022masked}, which applies $50\%$ random masking to the input data. 
The effect of masked autoencoder could be understood as data augmentation, which allows the encoder to focus more on the key information~\citep{xu2022masked}. 

\subsubsection{Results from the Bayesian discovery via SSGL prior}

We discuss both Laplace and Spike-and-slab priors for learning real moving rod data. 
In this case with very low and noisy data, the Laplace prior identifies an incorrect term $z$ over than the true target on $\sin(z)$. 
We provide detailed discussion of the Laplace prior in the Supplemental material (\ref{supp:laplace}). 
The SSGL prior could perform correct dynamical identification yet with a slightly large coefficient estimation. 
For the setting of SSGL prior, we set $\delta=0.09$, $v_0=1.0$, $v_1=5.0$, $\omega^{(k)}=0.01\times (0.999)^k$. 
The success discovery rate is $100\%$ from 10 training instances, which improves from $70 \%$ success discovery rate in the LASSO setting. 
Bayesian SINDy Autoencoder requires $1,500$ epochs for training. 
The best test error is $0.767$ for decoder loss, $0.305$ for the reconstruction of $\ddot{x}$, and $0.0263$ for the reconstruction of $\ddot{z}$. 
The fraction of the unexplained variance of decoder reconstruction is $0.003$.
For SINDy predictions, the fraction of unexplained variance of $\ddot{x}$ and $\ddot{z}$ reconstruction are $0.081$ and $0.961$. 

\begin{figure}
    \centering
    \begin{overpic}[width=\textwidth]{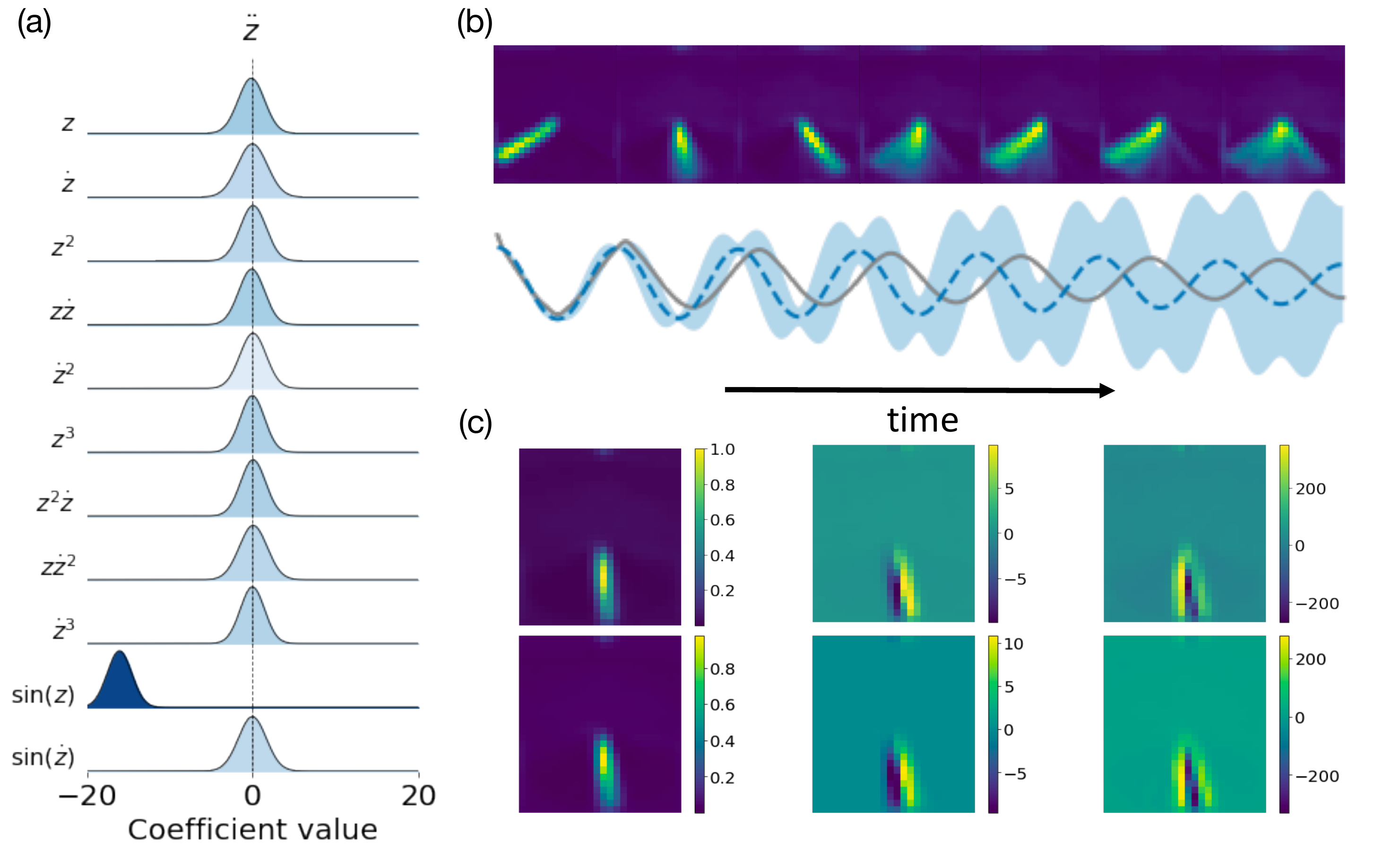}
    \put(31,41.0){\normalsize $z(t)$}
    \put(31,52.0){\normalsize $\hat{x}(t)$}
    \end{overpic}
    \caption{(a) Bayesian estimation and uncertainty quantification visualization of SINDy coefficient for real moving rod data under Spike-and-slab prior. (b) Visualization of prediction in both latent dynamics and pixel space with uncertainty quantification. (c) Reconstruction of Bayesian SINDy autoencoder for video frame, and temporal derivatives.  }
    \label{fig:ssgl_real_video}
\end{figure}

\begin{figure}
    \centering
    \includegraphics[width=\textwidth]{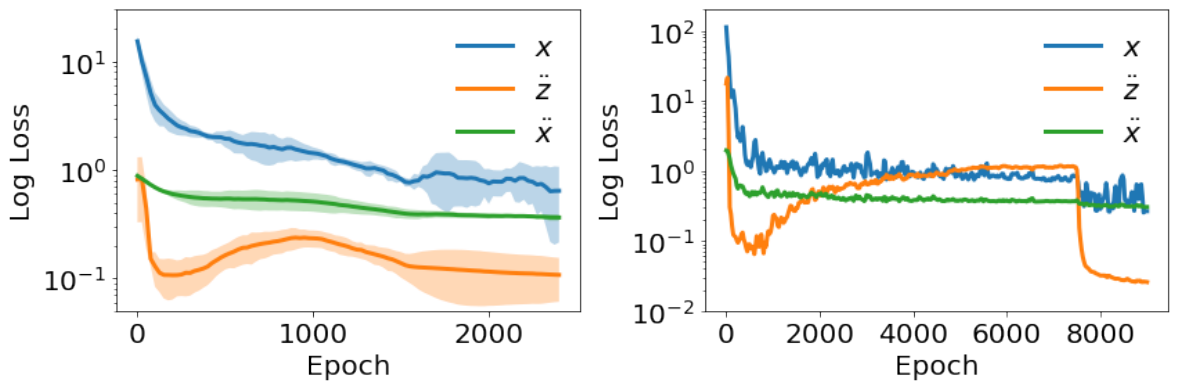}
    \caption{Left: by running the experiment 5 times with 2000 training epochs and 1500 refinement epochs, we plot the mean and standard deviation of testing loss under the log scale. Right: by running the experiment once with 7500 training epochs and 1500 refinement epochs, we plot the testing loss under the log scale.}
    \label{fig:loss_real_video}
\end{figure}

The posterior samples from Bayesian SINDy are shown in Fig.~\ref{fig:ssgl_real_video} (a), and the prediction of trajectory with uncertainty quantification of latent dynamics is shown in the bottom figure of Fig.~\ref{fig:ssgl_real_video} (b). The discovered governing equation from the SSGL prior is 
\begin{align}
    \ddot{z} = -16.06\sin(z).
\end{align}

The latent dimension $\ddot{z}$ with SINDy correctly identifies $\sin(z)$ as the active term and generates an uncertainty estimate. Utilizing the uncertainty estimation in the parameter space, we could perform uncertainty quantification on the prediction space as in Fig.~\ref{fig:ssgl_real_video} (b). 
The training outcome of Bayesian SINDy autoencoder are shown in Fig.~\ref{fig:ssgl_real_video},~\ref{fig:loss_real_video}. 
First, from Fig.~\ref{fig:ssgl_real_video} (c), we note that the reconstruction of the Bayesian SINDy autoencoder is very promising. The top line of figures is input data, and the second line of figures is generated by the autoencoder. Even if the scale is slightly different, it is convincing that the overall reconstructions on $x, \dot{x}$ and $\ddot{x}$ are very good. 
This can be suggested by Fig.~\ref{fig:loss_real_video} that the log loss converges with more training epochs. Throughout these experiments, we uniformly set training epochs to be $1,500$ and refinement epochs to be $1,500$. This decision is suggested by the observation in a longer run ($7,500$ training epochs and $1,500$ refinement epochs) that the log loss remains stable after 1000 epochs. For the reconstruction of $\ddot{z}$, having longer training epochs will harm the testing error due to overfitting. We note here the discovery by SSGL prior has dominant performance in testing loss comparing to SINDy autoencoder as shown in~\ref{supp:laplace}. This suggests the superiority of the SSGL prior in the real video data setting.

As suggested in Sec. 3.4, the Bayesian SINDy autoencoder enables a trustworthy solution in deep learning for video prediction. 
In Fig.~\ref{fig:ssgl_real_video} (b), we predict the following frames of video using the decoder and the simulation of latent dynamical systems. We observe that in the very beginning of future time frames, the prediction is very confident and accurate. Yet as we move to a longer time interval, the prediction becomes very uncertain, and we can observe this interesting phenomenon in Fig.~\ref{fig:ssgl_real_video} (b). 

\begin{figure}
    \centering
    \begin{overpic}[width=\textwidth]{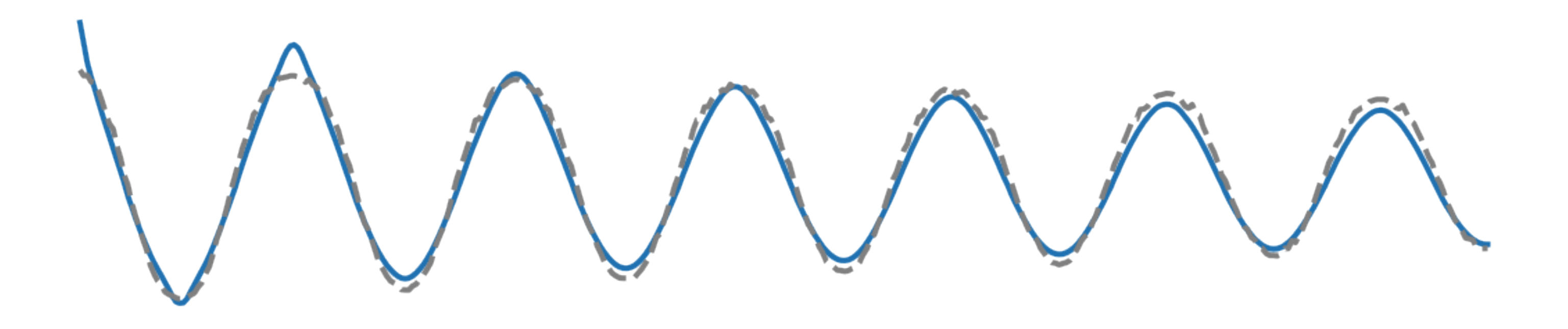}
    \linethickness{1pt}
    \put(5,1){{\color{black}\vector(1,0){20}}}
    \put(11,-1.5){\large Time ($t$)}
    \put(0,8){\LARGE $\theta(t)$}
    \end{overpic}
    \caption{Latent dimension visualization after rescaling (blue curve) versus manually labelled moving rod angle by human (grey curve). }
    \label{fig:real_video_latent_dim}
\end{figure}

It is essential to understand the latent dimension of the autoencoder in order to know the validity of the coordinate system discovery. Therefore, we create the following analysis on the latent dimension. We first manually label all angles $\theta$ from all video frames by humans, as shown in the grey dashed curve in Fig.~\ref{fig:real_video_latent_dim}.  
The blue curve represents the rescaled latent dimension $z$, and we observe these two curves match mostly perfect to each other. We apply rescaling in this process due to the coordinate system for the angles $\theta$ are equivalent after arbitrary transformation in its scale. For example, the angles represented by radians and degrees are equivalent to each other under different scales. 

\paragraph{On the discovery of the standard gravity constant $g$}

The SSGL prior accurately discovers the model with $\sin(z)$ term and reveals the standard gravity constant $g$. 
In our experimental setting, the length of the rod in our experiment is $123$ cm with the initial condition at $75^\circ$. 
We have an estimate of $g$ given these conditions that $\hat{g}=-9.876$. 
The estimation $\hat{g}$ is slightly overconfident and biased which concentrates at $(-9.889, -9.865)$. The bias is more likely to be removed with an enlarged video datasets. 
It is also possible to use the Laplace prior discovery to estimate the standard gravity constant because the discovered dynamical system with the term $z$ empirically has the form that $\ddot{z}=-\frac{\tilde{g}}{2L}z$ (c.f.~\citep{champion2019data} S2.3). This would result in an estimation of the standard gravity $g$ to be $\tilde{g}=-8.733$. However, the estimation $\tilde{g}$ is more biased, and lacks rigor in the context of the physics discovery.

Given this low-data and high-noise setting, the estimate from SSGL prior $\hat{g}$ is very remarkable given the true value that $g\approx -9.807$. 
Even if the discovery of the coordinate system could be up to an arbitrary scaling, the frequency of the dynamical system could uniquely identify the coefficient of $\sin(z)$. 
This validates the discovery under random scaling group transformation in the latent dimension. 

\section{Discussion}

In this paper, we design and implement a Bayesian SINDy autoencoder for automated coordinate and governing equation discovery from high-dimensional data. 
Using the Bayesian learning framework and sparsity-promoting priors, the proposed model identifies sparse dynamics in the latent dimension. 
Through experimental studies on synthetic Lorenz, a reaction-diffusion system, a pendulum, and real video of moving pendulum. We successfully perform model discovery under a learned coordinate system. In the small-data and high-noise regime for real video data, we identify the correct physical law with close estimation of the standard gravity constant $g$. 
Besides the model and physics discovery, in terms of video prediction, the Bayesian SINDy autoencoder provides uncertainty-aware future predictions with an exact understanding of its underlying physics, which enables a trustworthy alternative for deep learning-based video prediction. 

From our perspective, the Bayesian SINDy autoencoder has great potential to enable the concept of 'GoPro physics'~\citep{quanta_magazine_2022}, given the initial accomplishment in real video data discovery. However, the current framework does have several limitations, which hinder a more general application in various problems. 
The largest limitation comes from the current framework requiring a high manual workload on hyperparameter tuning. The Bayesian SINDy autoencoder has over ten hyperparameters, including the encoder-decoder size, loss function weights, and the sparsity-promoting prior. The current inferential process relies heavily on a reasonable hyper-parameter setting. 
There are several potential strategies to improve the hyper-parameter tuning part to improve the robustness of the training outcome. 
\begin{enumerate}
    \item The training process will be unstable and converge to a suboptimal solution if $\lambda_1, \lambda_2, \lambda_3$ is misspecified. To solve this problem, one can directly solve the multi-objective optimization for encoder-decoder architecture using methods like Multiple Gradient Descent Algorithm~\citep{sener2018multi}. 
    Such adaptations can not only work in this paper on Bayesian SINDy autoencoder, but will also be able to generalize into a wider context for physics-informed machine learning. 
    \item The hyperparameter setting for the sparsity level $\delta$, spike-and-slab distribution variances $\nu_0,\nu_1$, and environmental noise $\sigma$ can be tackled using a full Bayesian inference setup as described in~\citep{deng2019adaptive,rovckova2014emvs}. However, this will exceedingly enlarge the computational requirement and the difficulty in implementation. 
    \item Under minor misspecification of $\lambda_1, \lambda_2, \lambda_3$, one may consider to utilize an ensemble of trained Bayesian SINDy autoencoder initialized from various starting points~\citep{wilson2020bayesian,wu2021quantifying}. This process could ultimately construct a posterior distribution of the SINDy coefficient $\Xi$, and we can perform subset selection via stability selection or inclusion probability thresholding~\citep{fasel2022ensemble,meinshausen2010stability}. Another consideration would be applying variational autoencoders~\citep{kingma2013auto} to avoid the hard-thresholding procedure. Applying variational autoencoders could also help to perform better subset selection with Bayesian uncertainty estimation to the SINDy coefficient~\citep{kingma2013auto}. 
    \item The global minima exists, representing the real governing equation with the best coordinate system, but the nonconvex nature of neural network training makes it difficult to be found. The nonconvexity in neural network loss space is an open problem with very limited theoretical guarantees. To illustrate the existence of the global minima, from the case study for the moving rod (pendulum) video, the discovery with the $\sin(z)$ term leads to the minimal aggregated loss ($1.098$) throughout all experimental trails. All other suboptimal discoveries have a larger aggregated loss (for example via the Laplace prior, the discovery of $z$ has aggregated loss $1.559$). It is also observable from the data reconstruction (like in Fig.~\ref{fig:ssgl_real_video}) where the result from the Spike-and-slab prior is visually the best. However, there is no guarantee that the neural network training could always converge to the global minima. Again, applying the Bayesian Model Averaging initializing from various random initial conditions (as mentioned in the previous point) could be helpful, and this will be an interesting future direction. 
    \item The training of encoder-decoder structure requires fixing an approximately correct neural network depth and width. The current study on autoencoders still lacks a complete understanding of the structure. In general, an oversized encoder-decoder will result in an underdetermined system which is likely to result in an overfitted encoder-decoder. This will hinder the autoencoder from extracting the correct latent dimension as expected. Convolutional filters could be helpful in reducing the possibility of learning a wrong coordinate system by focusing more on the 2D features. 
\end{enumerate}

In conclusion, the Bayesian SINDy autoencoder is an improved option compared to the SINDy autoencoder especially in the small-data and high-noise limit. The Bayesian learning framework not only accelerates the training but also has dominating performance in various cases. 
Additionally, the Bayesian SINDy autoencoder enables uncertainty estimation, which is essential for data-driven decision-making purposes. Experimentally, we demonstrate the effectiveness of the Bayesian SINDy autoencoder using both synthetic and real video data. With only $14$ seconds of real moving pendulum video, the Bayesian SINDy autoencoder successfully discovers the coordinate system and governing equation and estimates the fundamental constant $g$. 
In future works, we hope to extend the current methodology into more complex scenarios with real video to explore further the power of Bayesian SINDy autoencoder in broad applications. 






\section*{Acknowledgments}

We acknowledge support from the National Science Foundation AI Institute in Dynamic Systems (grant number 2112085).  JNK further acknowledges support from the Air Force Office of Scientific Research (FA9550-19-1-0011).
We would also like to thank  Dr.~Wei Deng, Dr.~Bethany Lusch, Prof.~Simon Du, Prof.~Lexing Ying, and Dr.~Joseph Bakarji for their insightful suggestions and comments. 

\section*{Appendix: Real video discovery via the Laplace prior}
\label{supp:laplace}

We present the result from the Laplace prior which is suboptimal in this case. In this case with very low and noisy data, the Laplace prior identifies an incorrect term $z$ over than the true target on $\sin(z)$. 
However, the discovery on $z$ is an elegant failure due to the $z$ term also captures part of the dynamics via the identification on $z$.

For the setting of Laplace prior, we set $\lambda_3=1\times 10^{-2}$. 
The success discovery rate is $70\%$ from 10 training instances while the other $30\%$ instances will lead to zero discovery, indicating no active indices. 
Bayesian SINDy Autoencoder requires $1,500$ epochs for training. 
The best test error is $0.779$ for decoder loss, $0.364$ for the reconstruction of $\ddot{x}$, and $0.416$ for the reconstruction of $\ddot{z}$. 
The fraction of unexplained variance of decoder reconstruction is $0.012$.
For SINDy predictions, the fraction of unexplained variance of $\ddot{x}$ and $\ddot{z}$ reconstruction are $0.210$ and $0.941$. 

\begin{figure*}[ht]
    \centering
    \begin{overpic}[width=\textwidth]{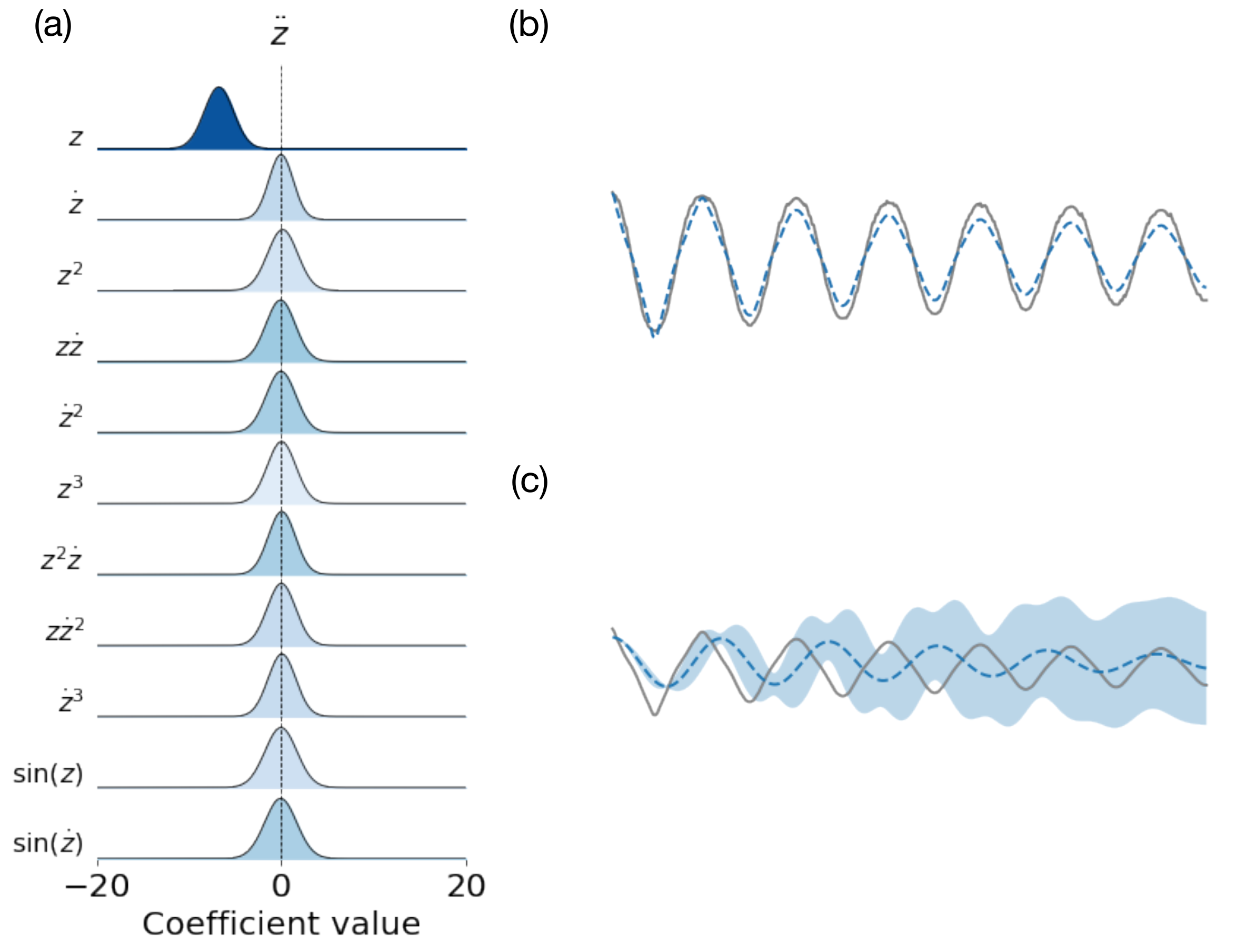}
    \linethickness{1pt}
    \put(49,16){{\color{black}\vector(1,0){30}}}
    \put(58,12){\LARGE Time ($t$)}
    \put(42,55){\LARGE $\theta(t)$}
    \put(42,22){\LARGE $z(t)$}
    \end{overpic}
    \caption{(a) Bayesian estimation and uncertainty quantification visualization of SINDy coefficient for real moving rod data under Laplace prior. (b) Latent dimension visualization after rescaling (blue curve) versus manually labelled moving rod angle by human (grey curve). (c) Visualization of predicted dynamics in latent space with uncertainty quantification. }
    \label{fig:laplace_real_video}
\end{figure*}
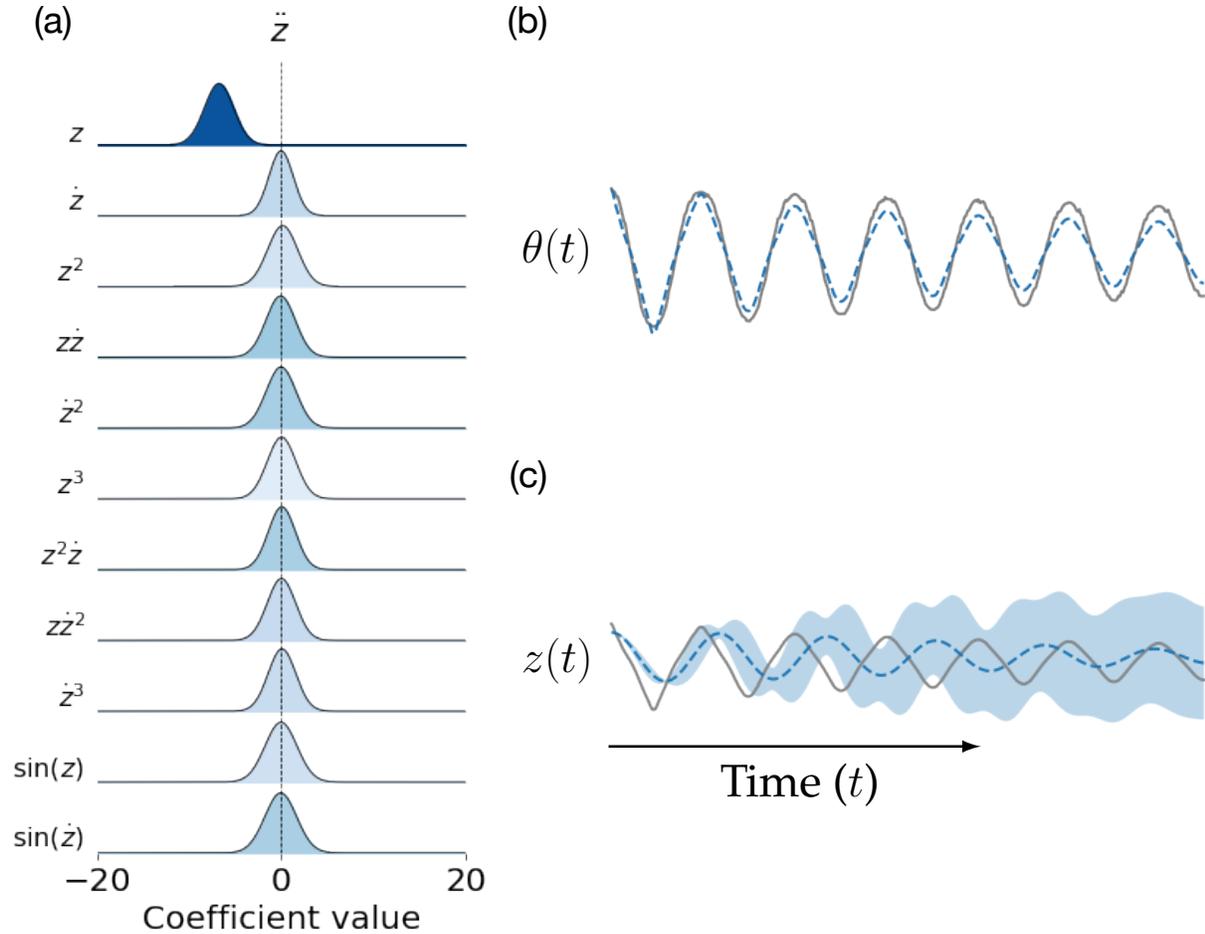

The posterior samples from Bayesian SINDy are shown in Fig.~\ref{fig:laplace_real_video} (a) and the prediction of trajectory with uncertainty quantification of latent dynamics is shown in Fig.~\ref{fig:laplace_real_video} (c). The discovered governing equation from the Laplace prior is 
\begin{align}
    \ddot{z} = -7.10 z.
\end{align}

The discovered governing equation with $z$ slightly deviates from the true dynamics similarly as reported in~\citep{champion2019data}, but still creates reasonable uncertainty estimate under model misspecification. 
The latent dimension in Fig.~\ref{fig:laplace_real_video} (b) follows a similar process in Fig.~\ref{fig:real_video_latent_dim}.  
The latent dimension $z$ still matches the human labels, but not as close as the learned result by the SSGL prior.

\bibliographystyle{plain}
\bibliography{main.bbl}




\end{document}